# Examining The Differential Risk from High-Level Artificial Intelligence and The Question of Control

Kyle A. Kilian[1], Christopher J. Ventura[2], and Mark M. Bailey[3*]

[1]National Intelligence University, USA; [2]Center for Intelligence in Extremis, National Intelligence University, USA; [3]Oettinger School of Science and Technology Intelligence, National Intelligence University, USA

Disclaimer: *All statements of fact, analysis, or opinion are the authors' and do not reflect the official policy or position of the National Intelligence University, the National Geospatial-Intelligence Agency, the Office of the Director of National Intelligence or any of its components, or the U.S. government.*

## Abstract

Artificial Intelligence (AI) is one of the most transformative technologies of the 21st century.  The extent and scope of future AI capabilities remain a key uncertainty, with widespread disagreement on timelines and potential impacts. As nations and technology companies race toward greater complexity and autonomy in AI systems, there are concerns over the extent of integration and oversight of opaque AI decision processes.  This is especially true in the subfield of machine learning (ML), where systems learn to optimize objectives without human assistance. Objectives can be imperfectly specified or executed in an unexpected or potentially harmful way. This becomes more concerning as systems increase in power and autonomy, where an abrupt capability jump could result in unexpected shifts in power dynamics or even catastrophic failures. This study presents a hierarchical complex systems framework to model AI risk and provide a template for alternative futures analysis. Survey data were collected from domain experts in the public and private sectors to classify AI impact and likelihood. The results show increased uncertainty over the powerful AI agent scenario, confidence in multiagent environments, and increased concern over AI alignment failures and influence-seeking behavior.

**Keywords**: Artificial Intelligence, Risk Analysis, Alternative Futures, Control Problem, AI Alignment, AI Governance

[*]Corresponding Author: mark.m.bailey@odni.gov

# 1. Background

Revolutionary discoveries in science and technology often happen in leaps. While science generally accumulates incrementally, some breakthroughs occur suddenly, delivering unprecedented opportunities and challenges. Following a lecture by renowned physicist Ernest Rutherford in 1933—where he dismissed the idea of nuclear fission as nothing but "moonshine"—younger scientist Leo Szilard, irritated by the remark, uncovered the key insight for the nuclear chain reaction within 24 hours (Jogalekar, 2013). Szilard's discovery upended the establishment and reshaped the future of science and geopolitics. Analogous discontinuities in science, or *technological transitions*, can emerge unpredictably, driving cascades of follow-on innovations, feedback loops, and higher-order changes (Valverde, 2016). Recent progress in artificial intelligence (AI) has sparked similar discussions of technological transitions, with some experts claiming it could be the "next electricity" (Lynch, 2017). Indeed, AI has enmeshed every aspect of society, increasing productivity, safety, scientific discovery, and economic well-being. However, as AI increases in complexity and capability, there are concerns about the scope of integration and autonomy. While largely beneficial in narrow domains—such as playing chess and recommending movies or products—AI used in autonomous weapons, for example, or safety-critical systems could pose significant risks (Maas, 2019).

AI systems have shown rapid improvements in capability and generality in recent years, most notably in the subfield of machine learning (ML). ML systems learn and adapt independently by drawing inferences from patterns in data without specified instructions. Independence, adaptation, and generalization—the ability to accurately transfer learned skills in one domain to new challenges—continue to demonstrate impressive results in AI systems (Hernández-Orallo, Sheng Loe, Cheke, Martínez-Plumed, & ÓhÉigeartaigh, 2021). While there are obvious differences and limitations, this general learning is what brings machine intelligence closer to human capabilities. Many milestones in capability and generality (once considered unattainable) are being met with increasing frequency. This acceleration is due in part to the sharp increase in data and computational resources but also to feedback loops of complementary technologies. Pending a significant disruption, the pace of AI technology is poised to continue to accelerate. A transformational breakthrough, with rapid capability gains and increased cognitive independence, could be a high-impact event with unpredictable consequences. What was once generally reserved for Hollywood science fiction has increasingly received serious debate. (Perry, 2020)

With the ability to learn and adapt without step-by-step instructions, advanced AI systems could promise novel solutions to once-intractable problems. However, this novelty can also lead to unexpected results or failures, what researcher Anca Dragan has termed "unexpected side effects" (Dragan, 2020, p. 134). While generally considered minor problems in contemporary AI applications (e.g., video games), the stakes could be significantly higher as systems increase in capability and generality. Indeed, researchers have voiced concerns about our capacity to control AI systems, with warnings of unanticipated failure modes that could lead to unintended military escalations and jeopardize command and control (C2) systems (Bailey, 2021) (Bailey & Kilian, 2022).

Analyzing this problem requires understanding the spectrum of risks associated with AI. The principal medium to long-term risks from advanced AI systems fall into four categories:

risks of misuse, accidents, agential, and structural. The *misuse* class includes elements such as the potential for cyber threat actors to execute exploits with greater speed and impact or generate disinformation (such as "deep fake" media) at accelerated rates and effectiveness (Buchanan, Bansemer, Cary, Lucas, & Musser, 2020). *Accidents* include unintended failure modes that, in principle, could be considered the fault of the system or the developer (Zwetsloot & Dafoe, 2019). *Agents*, or instantiations of AI that can interact with their environments and achieve goals autonomously, comprise most modern AI systems (e.g., AlphaGo). While there are multiple types of intelligent agents, goal-based, utility-maximizing, and learning agents are the primary concern and the focus of this research (Figure 1). AI agents learn much like humans in that they develop independent strategies to achieve objectives. Finally, *structural risks* are interwoven through social and technological systems—more complex and less easy to identify—requiring that we broaden the definition of risk. Structural risks are concerned with how AI technologies "shape and are shaped by the environments in which they are developed and deployed" (Zwetsloot & Dafoe, 2019). This risk classification schema is outlined in Figure 2.

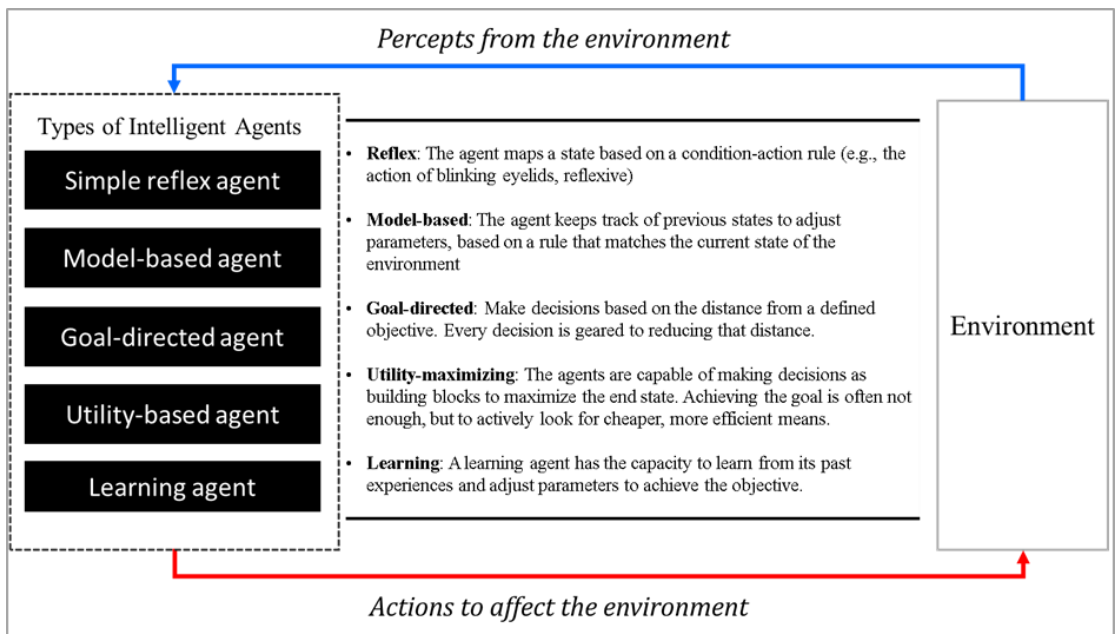

**Figure 1.** The five classes of intelligent agents. AI agents range in cognitive complexity from simple-reflex agents (e.g., automatic reaction given an environmental state change) to highly complex utility-based and learning agents. Goal-directed, utility-based, and learning agent types have the potential to exhibit agential characteristics with the risk of goal misalignment. (Russell, 2019)

| AI Risk Classification | | | | |
|---|---|---|---|---|
| | **Misuse** | **Accidents** | **Structural** | **Agential** |
| **Risk** | AI-enabled cyber attacks | Single system failures | Value erosion | Goal alignment failures |
| | Disinformation or misinformation | Multi-system failure cascades | Decision erosion | Inner alignment failures |
| | Deep fake media generation | Specification errors | Offense-defense balance disruption | Influence seeking |
| | Ubiquitous surveillance | Contagion and amplification | Uncertainty | Specification gaming and tampering |
| **Example** | *Fuzzing* attack | NYSE "Flash Crash" | Preference manipulation | Misaligned objectives |
| **Impact** | *Destructive* | *Catastrophic* | *Trans-generational* | *Existential* |

**Figure 2**: The spectrum of risk from advanced AI systems. The four classes are not necessarily confined to the degree of impact on the spectrum as there will be overlap across the spectrum (e.g., misuses and accidents can be catastrophic, and agential failures can be less dangerous). The spectrum is meant to provide a classification of the overall risk potential.[1]

One structural risk that is evaluated less frequently is the potential for automated systems to upend the stability of strategic weapons systems through the erosion of confidence. For example, alterations to behavioral regimes, such as nuclear rapprochement, can compromise trust and increase uncertainty (Shahar & Amadae, 2019, pp. 115-118). Beyond first-order risks (e.g., unintended weapons failures), researchers have noted that "higher-order" indirect risks from AI integration into peripheral systems (e.g., command and control, supply chain, etc.) can degrade multilateral trust in deterrence by increasing uncertainty in the trusted computing base of strategic systems (Shahar & Amadae, 2019, pp. 105-106). With respect to more direct agential risks, the potential for power-seeking and goal misalignment in connected agent systems could generate profound systemic risks and potentiate an unstable international system (Omohundro, 2008) (Carlsmith, 2022).

Since contemporary ML systems develop independent means to achieve objectives, problems with misinterpretation, unpredictability, or emergent subgoals are tangible concerns. For example, game-playing reinforcement learning (RL) agents are known to misinterpret objectives, game the system, or develop useful but unexpected subgoals to complete a task; agents tested in game environments have learned to prioritize the reward function (e.g., points) above the actual purpose of the game, resulting in "surprising, counterintuitive" failure modes (Dario, 2016). At the same time, AI's decision space seems to exist far outside the expected window of probability. An infamous example is AlphaGo's *move 37*, where DeepMind's system demonstrated its preternatural capabilities against world Go champion Le Sedol (Metz, 2016). While system failures and misuse capture most of the attention, there are subtle technical limitations and barriers to interpretation inherent in ML systems that will be far more difficult to resolve through governance and regulation.

---

[1] It is important to differentiate between outer alignment issues and inner alignment. Outer alignment refers to scenarios where an AI system interprets the human objective incorrectly or pursues it in a harmful way. However, the internal model could have separate misalignments. With inner alignment, the outer optimization process successfully internalizes the objective, but the inner model is itself an optimizer and goes about configuring the task in an unexpected, misaligned way. Evan Hubinger provides an example of human evolution. Human evolution (the outer model, in this case) has the goal of maximizing genetic fitness, while humans (the inner model) have a goal of maximizing pleasure or minimizing pain, not the explicit goal of genetic fitness. Thus, human objectives would be misaligned with the outer objective function of evolution. A learned model that is also an optimizer is known as a "mesa optimizer" and the objective, the "mesa objective."

This study investigates the risks and uncertainties that could arise from advanced AI development (including artificial general intelligence or AGI) and models how variations in social and technological change can impact outcomes. This research maps the four risk classes that most concerns AI researchers to plausible future scenarios, highlighting the variable impacts on international security. This work contributes primarily to the literature on future AI scenario development through a comprehensive AI risk framework. This paper also presents a novel exploratory modeling technique to characterize future scenarios and the associated risks with advanced systems. Through a hierarchical complex systems framework, the study structures the dimensions and key uncertainties of AI risk, such as technological transitions, competitive race dynamics, and control measures, to highlight the complex interdependencies and how they could yield highly variable futures.

## 2. Developing A Risk Framework

To develop a comprehensive risk framework, this research proceeds through a process of problem decomposition to isolate the most critical aspects of the AI ecosystem. Decomposition breaks down the components of a problem into smaller sub-problems, enabling the structuring of key uncertainties for scenario development. Generating scenarios typically involves identifying a set of influential drivers, which in combination create a range of plausible states (Virdee & Hughes, 2022). This framework expands on the process by developing a multitiered set of structural forces, influential drivers or dimensions, and plausible future states. Thus, the sub-components from problem decomposition are organized into three high-level classes (Figure 3), followed by 14 nested dimensions and 47 individual conditions (see *Appendix* for definitions).

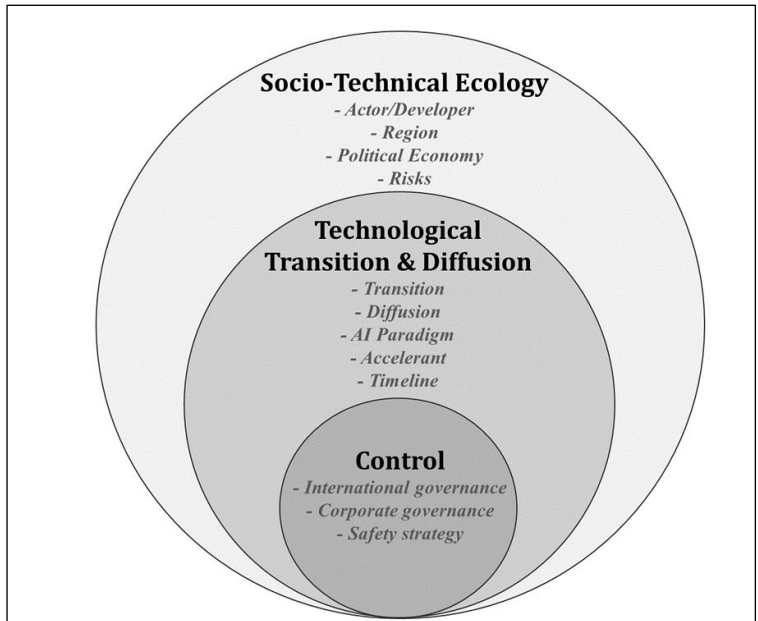

**Figure 3**. Advanced AI classification framework. The outer field includes macro-scale issues, such as actors, regions, and geopolitical race dynamics. The transition and diffusion subsystems include aspects of the technology itself that could directly impact system typology. Control is concerned with technical safety, institutional safety, and governance strategies.

The framework is structured as a hierarchical system of nested subsystems, dimensions, and potential conditions to analyze these processes as an ensemble of interdependent issues. The final system-level structure places the 14 individual dimensions as subordinate to the three higher-order classifications: technological transitions, socio-technical ecology, and control measures. There are six nested dimensions within the technological transitions class (Figure 4 and Figure 5), three within the socio-technical ecology (Figure 6), and five that are components of institutional and technical control (Figure 7). A level down the hierarchy, 47 conditions are compiled that represent plausible future states within each dimension.

### Technological Evolution, Transitions, and Diffusion

| Structural Forces/Dimensions | Definition | Conditions |
|---|---|---|
| **Capability & Generality** | Overall power of a system to achieve objectives, influence the world, and degree of generalizability across domains. | • **Low**: Systems remain approximately as capable as current systems.<br>• **Moderate**: Increase in cognitive capabilities and generalization across several domains.<br>• **AI Ecology**: Rapid capability gains across an ecology of decentralized agents; multiagent modular generalization.<br>• **AGI**: Systems can generalize to an approximate human-level parity and can recursively self-improve. |
| **Diffusion** | The distribution of systems at the time of development and the breadth of distribution; diffusion impacts the balance of power. | • **Decentralized**: system(s) are widely distributed with broad decentralized access and integration.<br>• **Multipolar**: system(s) developed and controlled by several leading institutions or governments.<br>• **Centralized**: system(s) developed by one research group or government program with restricted access. |
| **Technological Transition** | The rate of change of the system or technology as it increases in capability, including shifts in power dynamics and stability; e.g., the slope of phase transition. | • **Slow**: system(s) develop slowly, with linear progression (e.g., multiple decades).<br>• **Moderate uncontrolled**: systems develop with unexpected jumps in generalization (e.g., years/months/<decade)<br>• **Moderate controlled** (competitive): systems develop rapidly but are anticipated due to competition or conflict (e.g., years/months/<decade)<br>• **Fast**: A discovery leads to a discontinuous jump in generality, capable of recursive self-improvement (e.g., hours, days, weeks). |

**Figure 4**. AI Classification: Capability, Transition, and Diffusion. The three primary drivers of AI technological change: 1) system capability and generality, 2) diffusion and 3) rate of change/takeoff speed or technological transition. Each of the three can drive the velocity, character, and distribution of AI state changes, including variations in social response and control.

## AI Paradigm, Possible Accelerants & Timeline

| Structural Forces/Dimensions | Definition | Conditions |
|---|---|---|
| Paradigm | The paradigm or architecture that can achieve high-level capability and general-purpose functionality. | • **Current approach**: The current paradigms can scale to advanced capabilities and generalization (e.g., prosaic AGI)<br>• **New approach**: High-level generality requires a new paradigm or architecture<br>• **Hybrid approach**: Current paradigms are sufficient but require additional insights or a hybrid approach. |
| Accelerant | The technological insight or innovation that accelerates capabilities. | • **Compute overhang**: A new paradigm or slight modification can exploit existing computing more efficiently allowing rapid gains in generality.<br>• **Innovation**: A new insight or architecture, e.g., neuroscience or quantum computation, accelerates capabilities.<br>• **Embodiment** or data source: Technique for embodiment or a new quality of data provides rapid capability gains. |
| Timeline | The duration of the transition once a sufficient level of capability and generality is reached. | • **Less than 20 years**: AGI or a close approximation is developed before 2040.<br>• **From 20 to 40 years**: AGI or a close approximation is developed between 2040 and 2060<br>• **Greater than 40 years**: AGI or a close approximation is developed after 2060 years. |

**Figure 5**. AI Classification: Paradigm, Accelerants, and Timeline. The second group of dimensions in the transition cluster. Like the first three, paradigms can directly influence accelerant, transition, and timeline. Accelerants could be a disruptive innovation that influences the velocity of the system transition.

## Geopolitical Race Dynamics, Technical Ecology & Risk

| Structural Forces/Dimensions | Definition | Conditions |
|---|---|---|
| Race dynamics | The economic and geopolitical dynamics of increased competition and the effects on security, AI development timelines, and cooperation. | • **Cooperation**: AI is recognized as a global public good and cooperation increases.<br>• **Isolation**: Governments take a protectionist turn, causing wide disparities in cooperation, technical standards, and policies.<br>• **Monopolization**: Companies increase control over resources and influence, shifting the distribution of power.<br>• **AI Arms Race**: AI is named a strategic national asset and countries race for dominance. |
| Primary risk class | The highest impact risks from advanced AI systems, such as misuse, failures, and structural. | • **Misuse**: Cyber-attacks and disinformation increase in velocity and are the most disruptive risks.<br>• **Failures**: Systems are given more control over decision processes making agential risks and failures from misalignment more likely and dangerous.<br>• **Structural**: Increased decision autonomy brings subtle changes to society, human agency, and increased uncertainty. |
| Technical safety risk | The technical risks from agent systems that could pose a significant danger with advanced systems. | • **Goal alignment** (outer): Misaligned objectives remain the primary intractable problem with AI systems.<br>• **Power-seeking**: The most dangerous concern is the subtle acquisition of resources and runaway optimization processes.<br>• **Inner alignment** (mesa-optimization): Inner-misaligned failures and learned optimization present the most serious risks and are difficult to identify. |

**Figure 6**. AI Classification: Race Dynamics, Dominant Risk, and Technical Risk. The second cluster of nested AI dimensions is socio-technical ecology. Race dynamics impact the degree of competition, cooperation, balance-of-power, and dominant risks.

International Governance, Institutional & Technical Control

| Structural Forces/Dimensions | Definition | Conditions |
|---|---|---|
| AI Safety | Technical safety approach to align AI systems and their ability to transfer to more general-purpose advanced systems. | • **Scale invariant**: Current AI safety strategies are transferable to high-level systems regardless of scale.<br>• **New approach**: New AI safety approach must be developed from first principles to control a high-level instantiation.<br>• **Custom approach:** Each unique instantiation requires a specialized safety technique. |
| Actors | The entity that leads in the development of transformative advanced AI systems. | • **Coalition** (e.g., EU, NATO): Multinational alliances develop advanced AI.<br>• **Country**: A government develops the first advanced system.<br>• **Institution**: A private-sector organization (e.g., Tencent, Google) develops the first instantiation.<br>• **Individual**: An individual discovers advanced AI systems. |
| Region | The region that leads in advanced AI capabilities or develops the first instantiation. | • **USA-EU**: Actors in the US or the EU develop the first instantiation.<br>• **Asia-Pacific**: South, Southeast, Southwest, or East Asia develop the first high-level capability.<br>• **Other**: Central, South America, the Caribbean, or Africa make the discovery. |
| International Governance | The international governance bodies in place when advanced AI is developed. | • **Weak:** There is little improvement or a decline in AI governance.<br>• **Moderate**: There are modest improvements in international norms and institutions.<br>• **Strong**: International bodies and safety regimes are established (e.g., IAEA) along with multilateral agreements for misuse. |
| Corporate Governance | The degree of coordination on safety standard by AI companies. | • **Safety decrease**: There is a decrease in safety coordination across companies with an increase in zero-sum thinking.<br>• **Safety increase**: AI companies increase coordination and cooperation on technical safety standards.<br>• **Safety ideal**: AI research institutions increase coordination and agree on common safety standards. |

**Figure 7**. AI Classification: International Governance, Institutional and Technical Control. The third group of AI dimensions is control:1) Technical AI safety, 2) Actors, 3) Region, 4) International governance, and 5) Corporate Governance and safety standards. This set of dimensions represents the primary areas where change can be directly influenced through policy.

## 3. Data Collection & Methodology

This study implements an exploratory scenario modeling technique to understand the potential paths and risks of advanced AI development. The GMA method was designed to structure the total set of relationships that influence or interact with "multidimensional, non-quantifiable" problems (Ritchey, 2014) (Johansen, 2018). This study applies a variation of the alternative futures modeling framework devised by Blauvelt, et al. (Blauvelt, Jungdahl, & Closson, 2022) as a baseline for the clustering used for scenario development while concentrating on the complex interdependencies and the relationships between variables. The risk framework is used to populate the GMA matrix, with the dimensions and conditions arrayed across the model.

As outlined by Johansen, the GMA process can be described as a dialectical progression through repeated sequences of analysis and synthesis, with the matrix representing the entire morphological problem space (Johansen, 2018). GMA is a qualitative yet computational method where the value estimates for each condition provided by experts are computationally aggregated. Unlike the 2-dimensional risk matrix, with four possible solutions, GMA transposes the parameters into a multidimensional array with many plausible outcomes. Thus, the process reconceptualizes the matrix from the four-quadrant standard into a series of two-dimensional tables representing many dimensions. To structure the multidimensional matrix, the 14 dimensions and 47 conditions are arrayed across each axis of a Cartesian plane, with the dimensions and conditions duplicated along each for cross-reference and evaluation. The scenario combinations are reduced through cross-consistency assessment (CCA) to a smaller set of internally consistent configurations (Ritchey, 2014). The CCA step winnows down the prospective futures to a smaller set of configurations (removing logically inconsistent value pairs). This step provides a structured accounting of the morphological space, resulting in a relational database or a higher-order graph of connected typologies (Ritchey, 2014).

Once the conditions are structured across the matrix, each condition is cross-referenced and evaluated against every other (e.g., the condition *fast takeoff* with the condition *current paradigm*), averaging the values for impact and likelihood for each set of two conditions (Figure 8). Thus, each cell in the matrix is a computation between two conditions and four sets of values provided by survey participants. The configurations make up independent scenario combinations—or scenario pairs, two elements of a possible scenario (e.g., *AI Paradigm + Moderate takeoff*)—which are then clustered and categorized to drive the alternative futures analysis. Figure 8 depicts the transformation from the standard two-dimensional alternative futures matrix to an N-dimensional surface of interrelated tables (e.g., representing a standard 2-dimensional risk matrix for each parameter).[2]

---

[2] Taken from the standard 2-dimension matrix to 3, each box in the cube now represents the combination of three unique dimensions with impact and likelihood scores. Thus, one assessment cell, shown in Figure 8 as the dark ball, represents multiple sets of condition values for calculation with one potential scenario. The 3-dimensional matrix presented in Figure 8 is a simplified example to demonstrate the GMA process (the AI model is composed of 14 dimensions and 47 conditions and is difficult to illustrate).

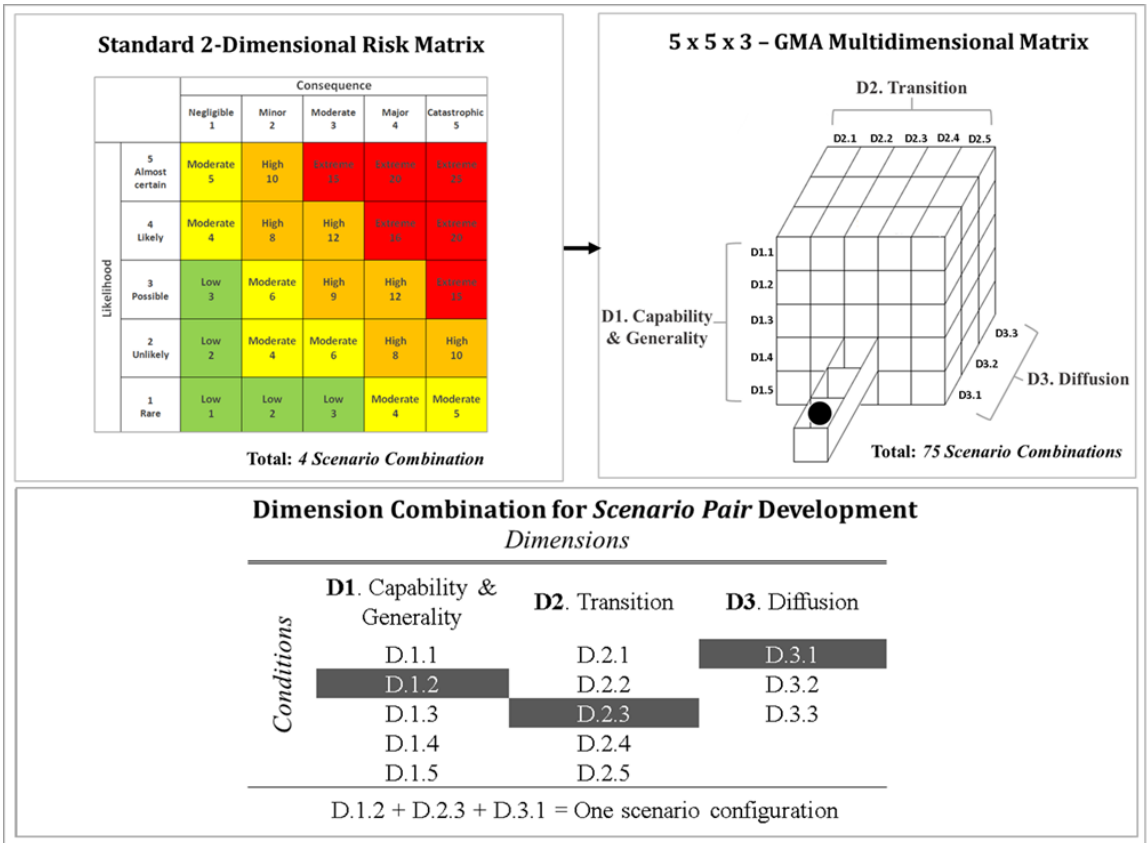

**Figure 8**. Standard four-quadrant impact-likelihood matrix vs the GMA multidimensional representation. (Ritchey, 2014) The standard two-dimensional matrix is typically used as the base for alternative futures analysis to arrive at four total scenarios (left) compared to the morphological problem space (right) that allows multidimensional structuring of relevant parameters. The dimension combination process (bottom) pairs conditions from each dimension to form one scenario configuration (D.1.2, D.2.3, D.3.1).

The number of simple configurations increases exponentially with an increase in the number of dimensions. For example, two dimensions with four conditions have one dyadic relationship (pairwise relationships between two dimensions), 16 paired value cells (assessment cells), and 16 pairwise relationships (4 conditions x 4 conditions); with an increase to four dimensions, there are now six dyadic relationships, 96 paired value cells, and 256 simple configurations (scenario pairs). As dimensions expand for highly complex problems, the possible scenario combinations increase sharply. With the 14 AI dimensions and 47 conditions, the number of potential future combinations is quite extensive, mapping as large a possible a space for AI scenario development. This is where GMA stands out as a method—the ability to systematically structure and evaluate as many features of a problem as needed for unique scenario configurations. For the AI model, the 47 conditions yield 15,116,544 unique combinations.[3]

With the interdependent relationships across dimensions, the GMA matrix can be modeled as a relational knowledge graph of system typologies using principles from network science (Ritchey, 2014). Indeed, the pairwise relationships between each condition can be represented as edges of a graph, with each dimension and condition acting as a node in the overall network. With distinct levels of higher-order classifications, subsystems, and conditions, graph network analysis can be a valuable tool to model complexity and is well suited to explore the intricacies and relationships of the AI model. Calculating a value pair in the CCA process is fundamentally the same as linking two nodes in the graph to evaluate the dyadic relationship. Thus, this research expands on the CCA method to evaluate the correlation coefficient and interdependency between each dimension and condition.

---

[3] The GMA dimension computations proceed as follows: two dimensions with four conditions: 4x4 (16 scenario pairs); four dimensions with four conditions: 4x4x4x4 (256 scenario pairs); the AI model with 14 dimensions and 47 conditions: 4x3x4x3x3x3x4x3x3x3x4x3x3x3 (15,116,544 scenario pairs).

For data collection, this study developed a comprehensive survey disseminated to domain experts to evaluate each dimension and condition (see *Appendix* for survey questions and definitions). The purpose of the survey was to establish a baseline of impact and likelihood assessment values for computation in the model. The questions were not, strictly speaking, questions but instead requests to judge or rank the impact and likelihood of each condition. While the GMA process generally uses qualitative decision support methods in a small group, this study incorporates a computer-aided variation to systematically structure the range of perspectives for scenario modeling. The surveyed experts assessed the impact and likelihood for each of the 47 conditions, providing two sets of values for each condition (Figure 9).

| Likelihood and Impact Scale ||||
|---|---|---|---|
| Likelihood || Impact on Security ||
| Very likely | 91-100% | Greatly decrease | 91-100 |
| Somewhat likely | 61-90% | Somewhat decrease | 61-90 |
| Even chance | 41-60% | Neither increase or decrease | 41-60 |
| Somewhat unlikely | 11-40% | Somewhat increase | 10-40 |
| Very unlikely | 0-10% | Greatly increase | 0-10 |

**Figure 9**. The likelihood rating scale is derived from the MITRE Corporations Risk Management Toolkit. (Engert & Landsdowne, 1999) The impact scale was developed independently to assess the security and stability ramifications of advanced AI.

After outlining the key dimensions and conditions that constitute the research problem and eliciting perspectives from experts, the GMA methodology proceeds through a five-step workflow (Figure 10): 1) collect and evaluate expert assessments, calculate descriptive statistics, and assemble the GMA matrix with the 14 dimensions and 47 conditions, arrayed across each axis of the matrix, 2) conduct a cross-consistency assessment (CCA) to remove logically inconsistent value pairs (e.g., duplicative self-referenced parameters) and average the combined impact and likelihood values for each condition, 3) evaluate the dimensions and conditions as a relational network (using Python's *NetworkX* library) (Hagberg, Schult, & Swart, 2008) to identify unexamined relationships, correlations, and sub-networks of tightly-connected parameters (e.g., community detection and maximum clique algorithms), 4) to cluster like values for scenario development, use the K-Means clustering algorithm to group values into scenario clusters for alternative futures (using Python's *ScI-Kit Learn* library) (Pedregosa, et al., 2011). The clusters are user-defined, allowing many tightly clustered arrays of values (e.g., 4 – 10) or only a few large groupings (e.g., 2 – 4) for in-depth world-building. Finally, 5) based on the interrelated dimensions and conditions, provide the exploratory scenario descriptions for synthesizing influential forces, interdependent factors, and differential risks.

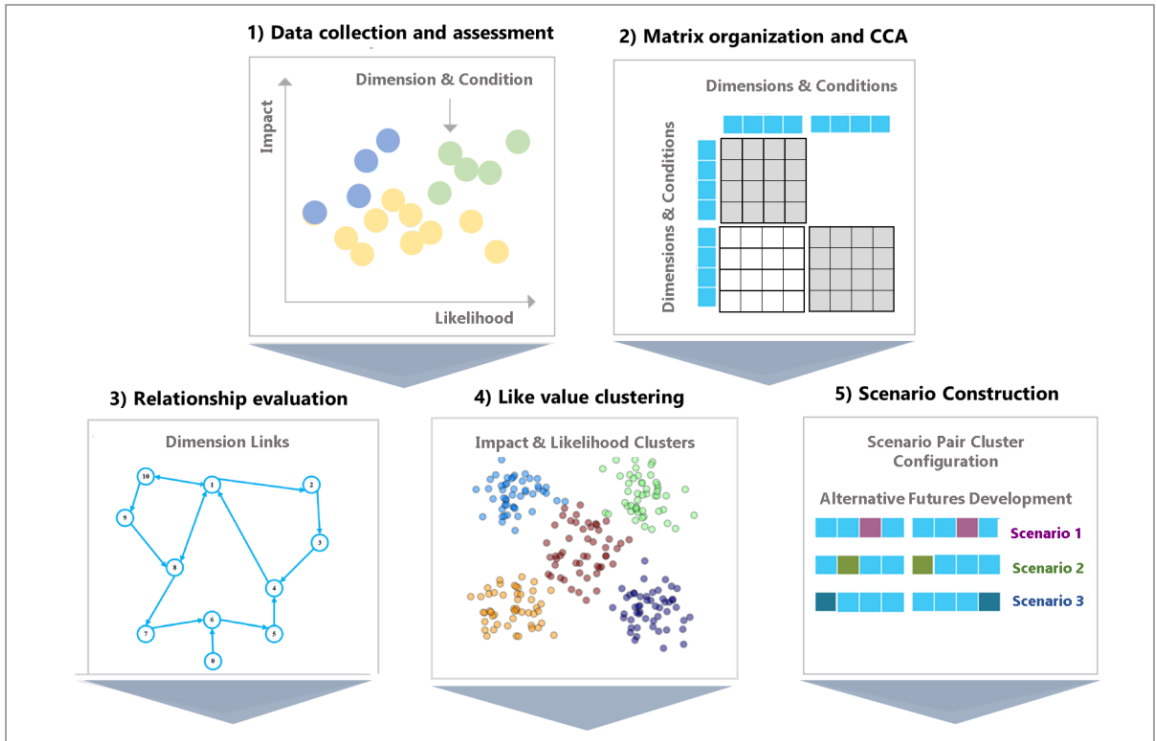

**Figure 10**. The graphic above displays the five-step workflow of this research methodology: 1) collect, organize, and evaluate data to structure the GMA matrix, 2) conduct a cross-consistency assessment (CCA), cross-evaluation, combination of scenario pairs, and computation, 3) evaluate the dimensions and conditions as a relational network, 4) cluster like values for scenario development, and 5) combine the completed scenario-pair clusters for exploratory scenario development.

# 4. AI Research Survey & Results

## 4.1 Survey Participation

The broad uncertainty (and disagreement) over developmental paths of advanced AI makes this topic especially difficult. In typical GMA studies, one group of researchers is gathered to provide their expert perspectives on an issue, which is suitable for established problems in policy or science where expertise is relatively abundant. However, with broad controversy, limited knowledge, and interdisciplinary issues to evaluate, the case is more complicated. Thus, for this work, a wide net was cast. First, a list was compiled of universities and research groups with dedicated programs on AI risk, major AI technology companies with AI safety teams, and existential risk organizations that focus on the issue. Responses were sought in public and private sectors. Several of the respondent organizations are within leading AI companies, including Deep Mind and Open AI. Second, a list of AI researchers actively working in machine learning was compiled, to narrow down respondents to those qualified to speak on the technical aspects of the problem. To broaden this base to include analysts that specialize in risk, the study identified analysts and organizations that focus on risks from emerging technologies and AI (including researchers in defense and intelligence). This step was important specifically for the impact questions, which require some working knowledge of geopolitical and governance issues.

To broaden participation, popular AI alignment forums and existential risk conferences were targeted for elicitation. While the organizations and academic conferences in this category vary considerably on their level of expertise, the study included several questions at the end to self-report on level of expertise. The respondents were categorized as having either *basic* knowledge (knowledgeable of the issues and the fundamentals), *intermediate* knowledge (knowledgeable on AI risk or governance) or expert knowledge (knowledgeable and actively working in AI alignment or risk). With the diversity of expertise across the breadth of issues, the self-report questions allowed us to differentiate responses from enthusiasts to experts and those working in AI safety and governance. See Figure 11 for breakdown of the level of expertise per question category.

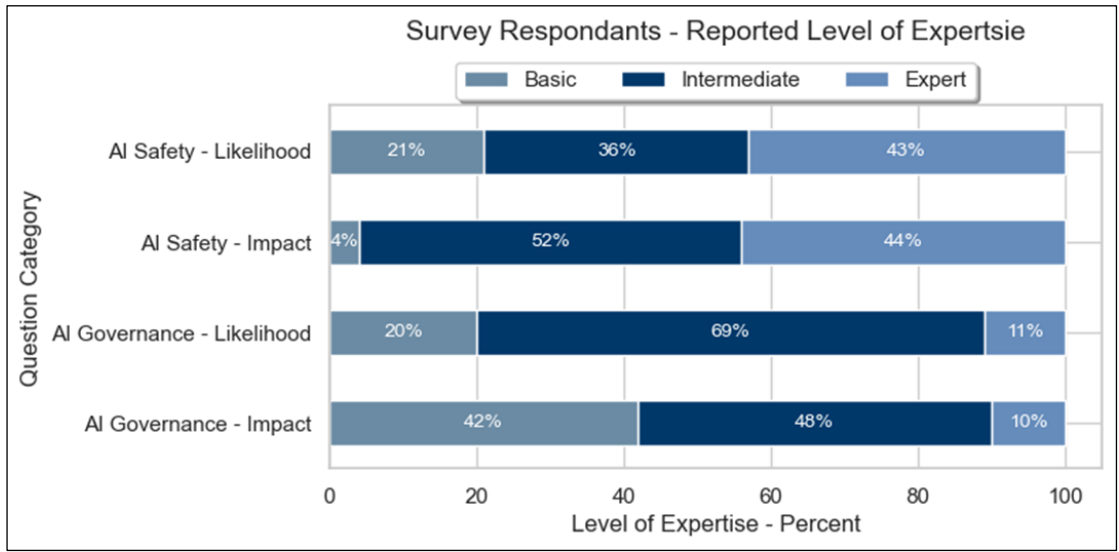

**Figure 11**: Survey Participation by reported level of expertise in AI safety and governance. This was devised to separate those with knowledge of AI alignment vs. knowledge of governance and to categorize level of expertise from basic (fundamental knowledge of the issue), intermediate (highly knowledgeable), and expert (knowledgeable and actively working in the field). This step was necessary given that not all alignment experts are knowledgeable on governance and vice versa.

From the range of AI practitioners, academics, data scientists, and existential risk scholars elicited for participation, the final tally resulted in 42 surveys for impact and 79 surveys for likelihood. The questions for impact were less intuitive for many experts to assess confidently, requiring substantive speculation and value judgments. This may explain the divergence in the completion rate for impact versus likelihood (user feedback highlighted this point). The quality of question type for impact requires intellectual flexibility, speculation, and comfort with exploratory "what if" reasoning with many assumptions (Marchau, Walker, Bloemen, & Popper, 2019). This style of exploratory analysis is more intuitive for practitioners in long-range risk analysis, policy (e.g., the RAND Corporation), and government but less so in scientific fields of study.

*4.2 Limitations*

There is a solid body of research that highlights the limitation in expert judgment, cognitive biases, and broad domain disagreements, which can be attributed to different world models and the appropriateness experts to judge potentially inappropriate issues (Cremer, 2021) (Armstrong, Sotala, & Ó hÉigeartaigh, 2014) (Granger, 2014). In addition, human judgment can be plagued by cognitive biases, such as confirmation bias (we seek to confirm beliefs, rather than refute them), ambiguity (our preference for options that are known), and the availability heuristic (judging recent events as more likely) (Kahneman, 2011) (Kay & King, 2020). Indeed, it is important for experts—and those using expert elicitation in research— to recognize the professional limitations in the face of radical uncertainty. An historical analysis of prior AI predictions shows a poor track record in technological forecasting while highlighting the need for intellectual humility (Armstrong, Sotala, & Ó hÉigeartaigh, 2014).

Notwithstanding the difficulties with prediction, novel exploratory models and decision support have proved especially useful for decision making given the complexity of the issue. Seth Baum and Anthony Barrett's fault tree analysis, Ross Gruetzemacher and Jess Whittlestone scenario mapping, and Shahar Avin's AI futures investigations are especially noteworthy examples (Baum & Barrett, 2017) (Avin, 2019) (Gruetzemacher & Paradice, 2019). In the spirit of Baum, Barrett, and Avin's research, this project does not aim to predict any specific technological path but rather to outline the breadth of plausible futures using exploratory scenario modeling; domain knowledge acts as the foundation, or baseline to orient our understanding of events and technological trajectories, given the observed conditions.

There are inherent limitations in the framing of this research and the breadth of the dimensions evaluated; for example, several of the dimensions are conditional (e.g., *paradigm* and *AI typology*) and are not all mutually exclusive, raising concerns with some participants. Thus, respondents were instructed to evaluate each condition independently as the interdependencies and conditional values were to be evaluated separately. At the same time, rather than a standard survey, this was designed as an exercise in ranking each condition on its impact and likelihood along a spectrum, where the best, middle, or worst option should be selected. Explained this way, rather than prediction, the participants were better able to understand and complete the process. The results suggest that the diversity of perspectives from public and private organizations, independent AI researchers, and emerging technology think tanks filled some of the gaps and limitations, resulting in a diverse set of perspectives for alternative AI futures.

## 4.3 High-Level Results

The study elicited judgments on the overall impact and likelihood of the 47 individual conditions from the risk framework. The respondents ranked the top high impact states to be a fast takeoff scenario (88) with AI arms race (86), influence-seeking (84) behavior, and a timeline under 20 years (78). Alternatively, participants ranked the lowest impact conditions as a slow takeoff scenario with strong corporate standards (12), international cooperation (13), and ideal governance (12). At the s time, the highest likelihood rankings include the US-EU developing advanced AI (78), with goal alignment a serious concern (74), and failure modes as the dominant risk (74), with advanced systems developed using a hybrid AI paradigm (72). The least likely conditions include a slow, gradual transition (33), with low capability and generality (27) developed by an individual (31) in a location other than the US-EU or Asia-Pacific region (0.27). The conditions listed in question order, with the average rankings are displayed in Figure 12.

| Technological Transition | | | | Socio-Technical Ecology | | | | Control | | | |
|---|---|---|---|---|---|---|---|---|---|---|---|
| Dimension | Condition | Impact | Likelihood | Dimension | Condition | Impact | Likelihood | Dimension | Condition | Impact | Likelihood |
| Capability-Generality | Low | 0.41 | 0.27 | Race dynamics | Cooperation | 0.13 | 0.37 | International Governance | Weak | 0.73 | 0.44 |
| | Moderate | 0.47 | 0.62 | | Isolation | 0.6 | 0.49 | | Moderate | 0.34 | 0.49 |
| | AI Ecology | 0.32 | 0.66 | | Monopolization | 0.71 | 0.63 | | Strong | 0.15 | 0.43 |
| | AGI | 0.74 | 0.51 | | Arms race | 0.86 | 0.67 | Corporate Governance | Decrease | 0.79 | 0.39 |
| Diffusion | Decentralized | 0.4 | 0.49 | Dominant risk | Misuse | 0.71 | 0.62 | | Increase | 0.29 | 0.54 |
| | Multipolar | 0.56 | 0.66 | | Failure | 0.68 | 0.74 | | Ideal | 0.12 | 0.49 |
| | Centralized | 0.75 | 0.38 | | Structural | 0.58 | 0.61 | AI Safety | Scalable | 0.34 | 0.42 |
| Transition | Slow | 0.2 | 0.33 | Technical safety risks | Outer | 0.7 | 0.74 | | New | 0.55 | 0.69 |
| | Moderate | 0.6 | 0.55 | | Influence | 0.84 | 0.65 | | Custom | 0.76 | 0.53 |
| | Competitive | 0.5 | 0.55 | | Inner | 0.56 | 0.57 | | | | |
| | Fast | 0.88 | 0.45 | Actor | Coalition | 0.15 | 0.38 | | | | |
| Paradigm | Current | 0.76 | 0.5 | | Nation | 0.39 | 0.6 | | | | |
| | New | 0.34 | 0.6 | | Institution | 0.77 | 0.7 | | | | |
| | Hybrid | 0.42 | 0.72 | | Individual | 0.56 | 0.31 | | | | |
| Accelerant | Overhang | 0.71 | 0.56 | Region | US-EU | 0.2 | 0.78 | | | | |
| | Insight | 0.3 | 0.71 | | Asia-Pacific | 0.66 | 0.54 | | | | |
| | Embodiment | 0.47 | 0.54 | | Other | 0.53 | 0.27 | | | | |
| Timeline | Under 20 | 0.78 | 0.42 | | | | | | | | |
| | 20-40 | 0.59 | 0.66 | | | | | | | | |
| | Over 40 | 0.34 | 0.43 | | | | | | | | |

**Figure 12**. The impact and likelihood rankings for the conditions of the AI model. Conditions are grouped hierarchically under dimensions and subsystems classifications.

While all organizations and individuals selected for participation in this study were qualified, understanding their areas of expertise and their degree of experience across the different fields provides some insight into how conditions were ranked. One difference that stands out is the distribution of rankings. Respondents new to the field, with a basic understanding of AI safety, ranked the conditions somewhat haphazardly, with strong variance across questions (Figure 13). This same pattern held for impact. On the other hand, intermediate and expert participants tended to trend closer to the mean, with much less variability, and with relatively consistent trend lines (e.g., expert and intermediate participants score AGI as high impact, with only variations in degree). Much of these patterns are likely due to level of experience working the issue; for those new to the subject matter, extreme values can seem more intuitive and appealing, while experience privileges more moderate, nuanced judgment.

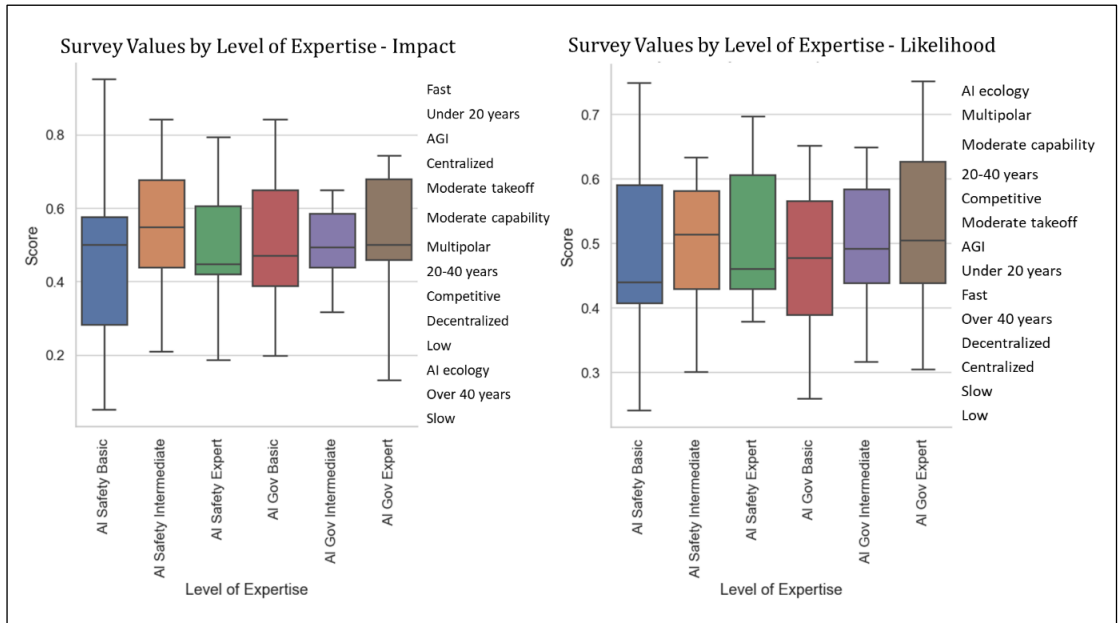

**Figure 13**. Box plot representing the variability among basic, intermediate, and expert levels of expertise for both AI safety and governance. Basic level expertise shows wide ranging scoring patterns and mild disagreement with intermediate and expert practitioners.

The dimensions that underpin AI technological change directly—the technological transitions cluster—are especially relevant to understand the overall trajectory of advanced AI. Indeed, the timeline, transition, and diffusion dimensions are all tightly interrelated and depend very much on the outcome of the others. Timeline requested that participants provide their best judgments on the approximate period where high-level general AI agents will be developed that can complete a substantive share of cognitive tasks. Transitions—fast, moderate, controlled, uncontrolled, or slow—reflect the velocity of intelligence progression (e.g., continuous, discontinuous, or moderate transition), the generality of systems, and the degree of power displacement. At the same time, diffusion has implications for the breadth of power distribution, system integration, and overall impact. Combined, the three dimensions outline the boundary of critical transitions (Figure 14). Transitions, timelines, and diffusion also directly influence and are influenced by economic and geopolitical race dynamics: a vicious circle of economic and geopolitical feedback mechanisms.

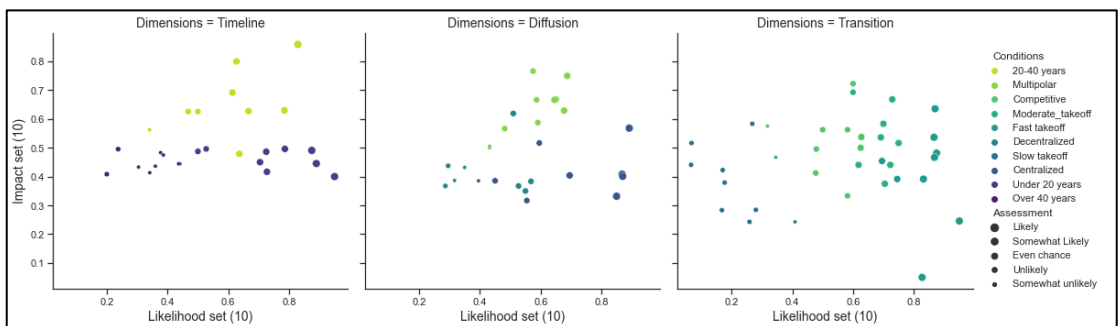

**Figure 14**. The impact and likelihood averages for technological transitions, diffusion, and timeline. Figure 14.1 (left) shows the impact and likelihood averages for the timeline dimension, 14.2 (center) the averages for the diffusion dimension, and 14.3 (right) the averages for the transition dimension. The individual conditions are color-coded, and the size of icon varies by reported likelihood. For legibility, the rankings are displayed as bins of 10 responses.

The participants ranked the most likely timeline of arrival from 1) 20-40 years at 66 percent, 2) Over 40 years at 43 percent, and 3) under 20 years at 42 percent. The spread between 20-40 and over 40 was somewhat surprising since the standard assessment over the past few decades approximated the upper limit of 40-plus years for most surveys (Date Weakly General AI is Publicly Known, 2022) (AI Impacts Staff, n.d.) (Barnett, 2020). For transitions, which speaks to capability jumps and rate of progress, participants assigned a moderate *uncontrolled* takeoff as a tie with moderate *competitive* at 55 percent. A moderate uncontrolled takeoff is broadly defined as an unexpected rapid increase in capabilities, likely too fast for society and institutions to adjust. In contrast, a moderate competitive takeoff is presented as a case where advanced general AI is actively pursued by governments and technology companies where rapid change is anticipated.

A core aspect of the moderate competitive takeoff condition is the control of resources and the militarization of technology. The uncontrolled moderate takeoff can be framed as an unexpected rising tide of intelligence generalization with rapid diffusion. In addition, by a relatively wide margin, participants gave multipolar diffusion a likelihood score of 66 percent, trailed by decentralized at 49 percent and centralized at 38 percent (Figure 15). Altogether, the timeline results from this research align modestly well with prior work but show an increase in expectations for powerful AI systems (AI Impacts Staff, n.d.). Checking the results against similar research on trends and attitudes, there is indeed a change in expectations and optimism for general AI that is likely due to the increasing evidence that universal scaling laws apply to computational capacity and can lead to unexpected capability gains (e.g., Metaculus general AI forecasts) (Grace, Salvatier, Dafoe, Zhang, & Evans, 2018).

Technological Transition, Diffusion, and Timeline Assessments

| Dimension | Condition | Impact Average | Likelihood Average |
|---|---|---|---|
| **Timeline** | **20-40 years** | **0.59** | **0.66** |
|  | Over 40 years | 0.34 | 0.43 |
|  | Under 20 years | 0.78 | 0.42 |
| **Diffusion** | **Multipolar** | **0.56** | **0.66** |
|  | Decentralized | 0.4 | 0.49 |
|  | Centralized | 0.75 | 0.38 |
| **Transition** | **Moderate takeoff** | **0.6** | **0.55** |
|  | **Competitive takeoff** | **0.5** | **0.55** |
|  | Fast takeoff | 0.88 | 0.45 |
|  | Slow takeoff | 0.2 | 0.33 |

**Figure 15.** Participant Assessments for Transition, Timeline, and Diffusion. The highest participant ranking for timeline is 20-40 years at almost 70 percent probability.

In 2022, the AI community has witnessed a handful of milestones in generalization that surpassed expectations, such as Google's Pathways model (PaLM), which can explain its reasoning across multiple tasks (chain of thought prompting), or OpenAI's DALL-E 2, which paints original art prompted by a textual or verbal description (TRTWorld Staff, 2022). In May 2022, DeepMind released its GATO transformer model in a paper titled "a generalist agent," which can generalize across 604 individual domains and make independent decisions based on context, whether to "play Atari, caption images, chat, stack blocks with a real robot arm and much more" without changing the model parameters (Reed, et al., 2022) (Grossman, 2022). While some experts have pushed back on the hype surrounding the release of PaLM, DALL-E, and GATO, many researchers have adjusted their timeline assessments for the emergence of general AI by decades; AI researcher Conor

Leahy claimed his expectations for AGI are now at "20% to 30% in the next five years. 50% by 2030, 99% by 2100, 1% had already happened" (Leahy, 2022).

The other key condition that scored lower than anticipated was *fast takeoff*, at 45 percent. The fast takeoff scenario underpins the standard AGI risk scenario as outlined by Yudkowsky and Bostrom (Yudkowsky, 2013) (Bostrom, 2014). A comparable condition to fast takeoff is *centralized* diffusion, where an advanced AI is discovered by a small group or solitary system and is maintained in a protected program. Participants scored centralized diffusion at 38 percent. Unsurprisingly, the *slow takeoff* condition was scored far lower than the others at 33 percent, echoing the increased expectations for more powerful systems. In addition, a *decentralized* diffusion, where advanced AI is discovered and widely accessible, was scored at 49 percent, at the approximate midpoint between multipolar on the high end (66 percent) and centralized on the low (38 percent). Combined, the low likelihood assessments for a fast takeoff and centralized diffusion point to a dynamic where researchers are inchingly finding flaws in the standard AI risk arguments. Indeed, there has been a growing debate over the empirical foundations of the standard AI transition model. For example, researchers Ben Garfinkel, Paul Christiano, and David Hanson have taken alternative stances on the slope of AI takeoff and technological diffusion, broadening the range of perspectives on future AI paths (Lempel, Wiblin, & Harris, 2020) (Chrstiano, 2019). Ultimately, the expansion of the AI scenario space has influenced new research directions across the AI safety and governance community.

*4.4 Parameter Correlations*

Correlations between two or more parameters in the AI futures model provide some insight into the multidimensional relationships. Correlations of technological trends have been used for years to understand relationships in technological innovation. Indeed, trends in precursor technology can be correlated to follow-on innovations at the cross-over point (e.g., transition), lending insight into the velocity of change and environmental factors (Martin, 1970). This study calculates the correlation coefficients for the higher-order dimensions as a starting point in the cross-reference analysis to evaluate interdependence. Since the raw assessment values are based on the conditions exclusively, the condition scores subordinate to each dimension are averaged to evaluate impact and likelihood at the dimension level (e.g., the average of decentralized, moderate, and centralized for the diffusion average). Most of the dimensions show high correlation values, with the majority displaying positive correlations. However, negative correlations are also observed, especially between impact and likelihood values of similar related classes (Figure 16). Dimensions reasonably expected to trend in concert, such as transition, accelerant, and timeline, show high correlation values overall but less so with some, such as AI paradigm. Dimensions that intuitively could have strong bidirectional influence or impact showed strong negative correlation values, such as risk and governance.

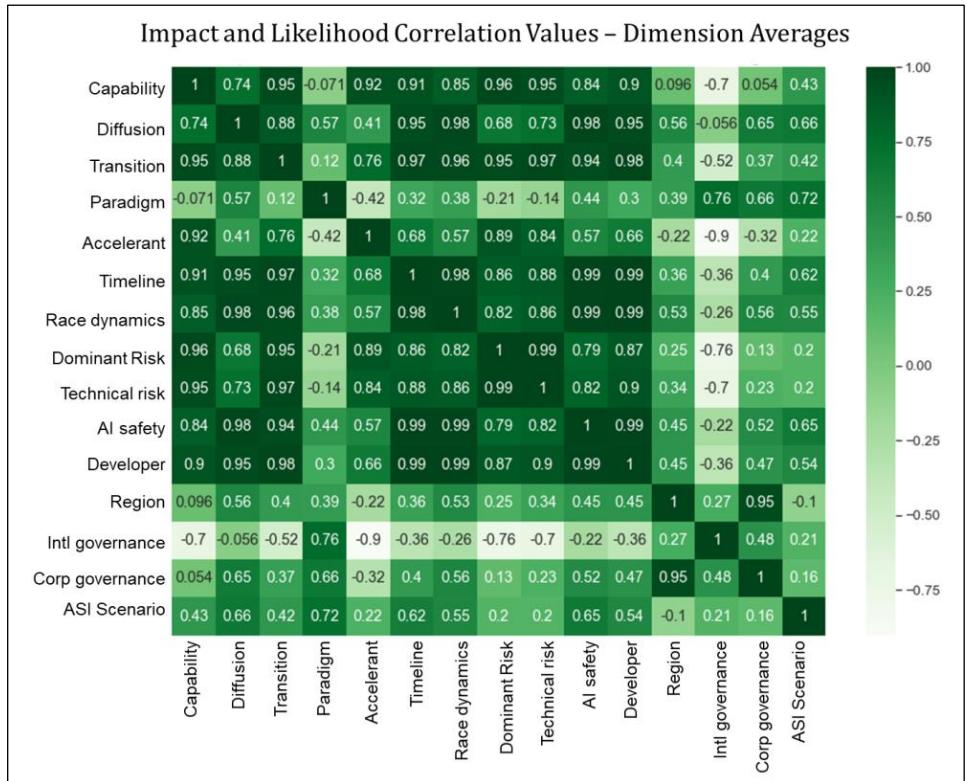

**Figure 16**: Correlation heatmap of impact and likelihood values for the dimensions. The darker green represents positive correlations and the light negative. The diagonal values are self-referenced and should be ignored.

The dominant and technical risk dimensions both show the highest positive correlation values (0.994). The top five positive correlations following risk include race dynamics and developer (0.993), timeline and developer (0.993), and race dynamics and safety technique (0.991). Additional high positive correlations that are relatively intuitive include distribution and race dynamics (0.97), transition and technical risk (0.96), capability and risk (0.96), and safety techniques and distribution (0.98). Two notable pairs include capability and technical risk (0.94) and capability and safety techniques (0.84) since the overall power of the system impacts the risk of misalignment and the safety strategies required. Additionally, the strongest negative correlations include accelerant and international governance (-0.9), capability and governance (-0.7), and risk and governance (-0.76).

At the level of conditions, the correlations between states show a higher degree of interdependence than between the dimensions independently. While the dimensions display the higher-order relationships, the interdependencies between conditions suggest that a change in one condition (e.g., *compute overhang*) could influence the fate of the other (e.g., *AGI capability*). The three key dimensions that could affect transition velocity, diffusion, and character, highlighted above as critical factors in AI timelines, show strong positive and negative correlations, largely with intuitively opposing trends (e.g., *fast takeoff* shows a strong negative correlation with the *over 40-year* timeline). The correlations between the three dimensions and conditions from the technological transition cluster are shown in Figure 17. There are some examples of weaker or less intuitive correlations (e.g., *multipolar*, *20-40 years*), but overall, the relationships hold.

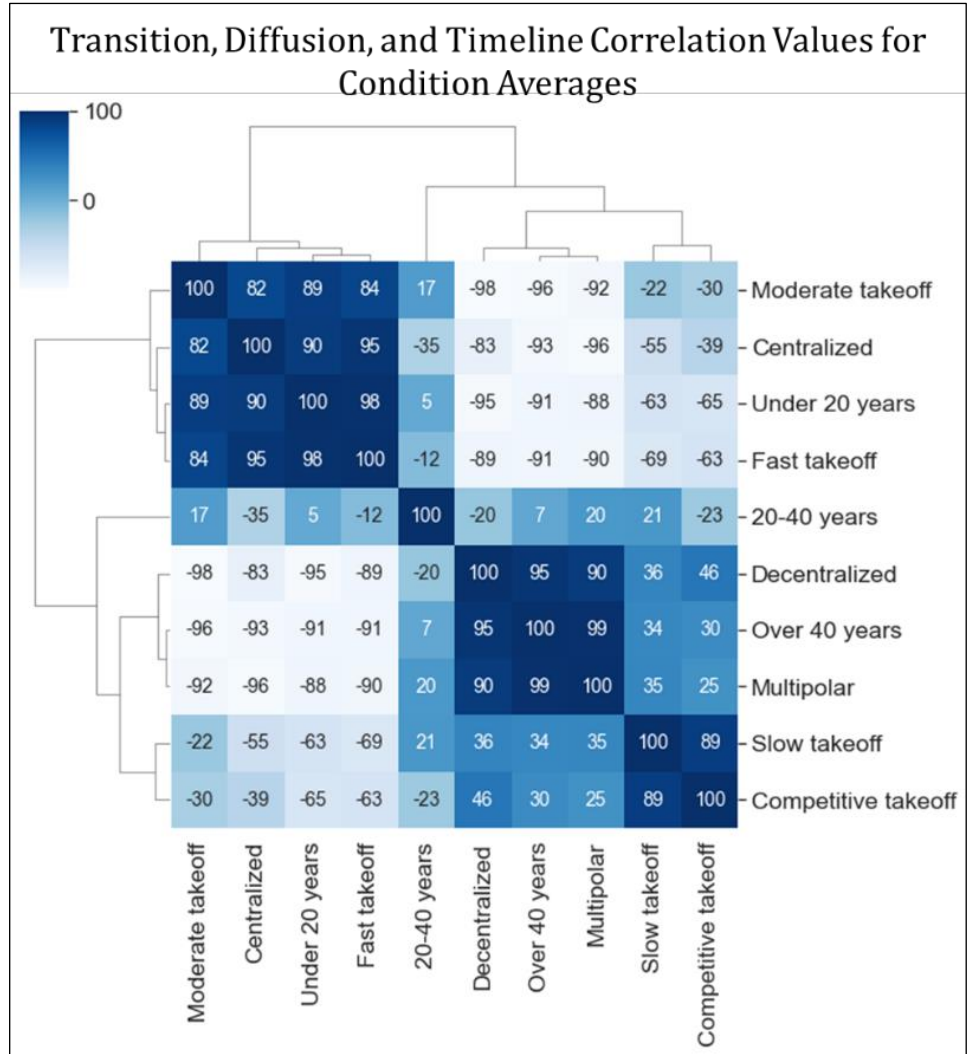

**Figure 17**. Correlation values between the condition states for transition, diffusion, and timeline. The strong interactions within the subsystem, either positive or negative, demonstrate how nested interactions could influence the direction of adjacent conditions.

The correlation matrices demonstrate the relationships between AI dimensions and conditions and highlight how differential development paths could lead to variable system typologies and technical and social risks. Understanding the relationships between dimensions and conditions could potentially act as a foundation for developing indicators for anticipatory analysis and system-level monitoring. At the same time, technological systems are too complex for reliable prediction (with insufficient knowledge, data, or initial conditions), understanding the dynamics of influential variables could be an important step. Indeed, the interplay between technologies, paradigms, and social dynamics could accelerate the trajectory of AI systems. Enabling technologies like quantum computing could trigger an acceleration in race dynamics, while a broadly distributed breakthrough has the potential to shift power dynamics away from countries and companies to populations (Huang, et al., 2021) (Kalluri, 2020).

To separate the most tightly linked interdependencies with the highest correlation values, a network graph is constructed and subset to the maximum connected components (Figure 18). As with the correlation matrices, the strongest linked dimensions remain between transitions, diffusion, paradigm, risks, and race dynamics. There is a consistent grouping of eight to ten influential conditions that show disproportionate influence on the others and the ultimate trajectory of the system. For a more detailed analysis, the raw impact and likelihood scores are graphed for all condition values. Displaying the graph for all connections above 0.6 correlation highlights two tightly woven clusters connected by four conditions: failure modes, goal alignment, mesa optimization, and the 20-40 years condition (Figure 19). These connections are intuitive given the centrality of technical and social risk to AI safety and the strength of the belief that general AI will emerge within 20-40 years.

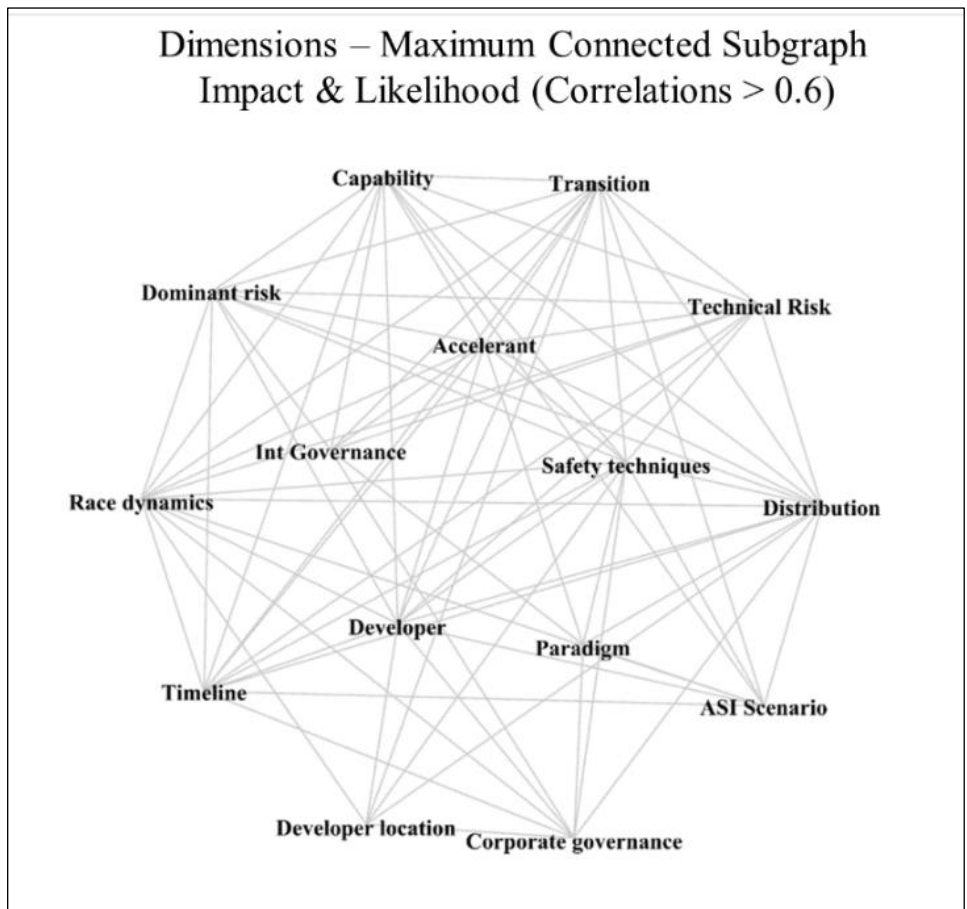

**Figure 18**. Connected Subgraph of AI Dimensions. The graph highlights all connected dimensions with over 60 percent correlation values. The dimensions with the highest concentration of connections are the strong positive correlations between accelerant, risk, transition, capability, developer, and diffusion.

Analogous to the high correlations and high impact and likelihood values, the two most common highly connected scores were in the risk class or technical risk categories. Thus, it lines up that failure modes in general, plus goal and inner alignment (or mesa optimization), would show strong betweenness centrality among the three clusters. The 20-40 years judgment for general AI displayed similar high centrality, as did risk.

Betweenness centrality is a metric of the importance of an entity to the flow of information between one segment of the network to another (e.g., the shortest path) (Golbeck, 2015).

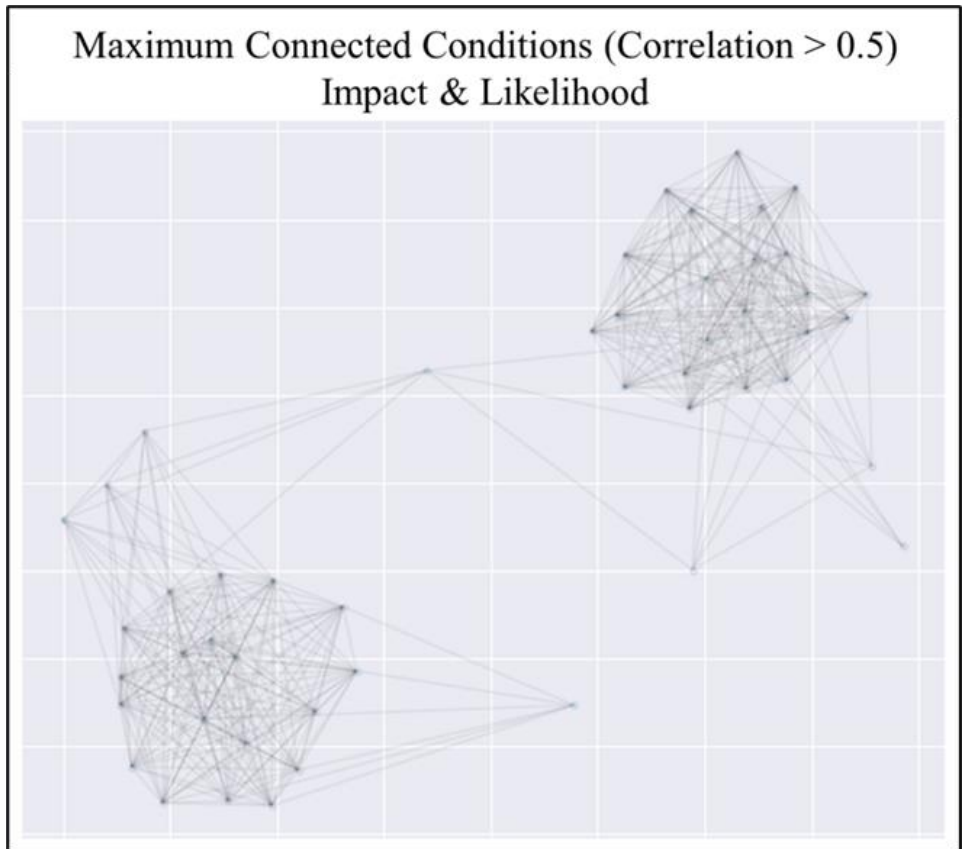

**Figure 19.** AI Communities Subgraph. There are several communities below correlation values of 0.7 that are linked by failure modes, goal alignment, mesa optimization, and 20-40 years. This is intuitive given the centrality of technical and social risk to AI safety and the strength of the belief that general AI will emerge within 20-40 years.

Exploring the network dynamics further to understand and measure these clusters, a community detection algorithm was performed to identify distinct groupings that share complex similarities. Using the *greedy modularity* algorithm, the two distinct communities were identified in the data (Hagberg, Schult, & Swart, 2008). Greedy modularity maximization starts with a group of entities in a given community and joins the next group that most increases *modularity* until no other pairs are present. It's a technique used to structure groups of values with given characteristics. Of the two communities identified, one contains a set of values that trend toward slower takeoff scenarios or distributed outcomes, while the other leaned toward centralized diffusion and fast takeoff (Al-Mukhtar & Al-Shamery, 2018). To go a level deeper, the "find cliques" algorithm is used to identify subsets of the two communities that make up even smaller subgraphs and groupings, known as the "clique problem" in networks science; a "maximal clique" is a subgraph where at least one of the vertices is in the larger set (Chang, Klok, & Lee, 2001) (Bron & Kerbosch, 1973). Thus, within each community network, there are smaller sub-networks that are more closely tied.

| Modularity Community Detection - Maximum Clique ||||
|---|---|---|---|
| **Clique 1** | **No. Clique Membership** | **Clique 2** | **No. Clique Membership** |
| Weak governance | 4 | Hybrid paradigm | 5 |
| Influence seeking | 4 | Safety ideal | 4 |
| Institution | 4 | Country | 4 |
| Under 20 years | 4 | Moderate governance | 4 |
| Arms race | 4 | New paradigm | 4 |
| Safety decrease | 4 | Moderate safety | 4 |
| Fast takeoff | 4 | Insight | 4 |
| Overhang | 4 | Coalition | 5 |
| Asia-Pacific | 2 | Cooperation | 4 |
| ASI 3 | 2 | USA-Western Europe | 4 |
| Custom techniques | 2 | Distributed ASI | 4 |
| Other | 2 | Strong governance | 4 |
| Current paradigm | 2 | Over 40 years | 3 |
| AGI | 2 | Current techniques | 3 |
| ASI 2 | 2 | Multipolar | 3 |
| Individual | 2 | Decentralized | 3 |
| Moderate takeoff | 2 | Moderate capability | 5 |
| Centralized diffusion | 3 | General techniques | 3 |

There are 14 distinct "cliques" observable in the data and 18 maximum cliques (Figure 20). These numbers seem unsurprising given that 14 dimensions could naturally align with 14 cliques; however, since each condition is not mutually exclusive, interact, and overlaps in a variety of ways, one would expect there to be variations across groups with different boundaries. Indeed, analysis of the cliques and community data demonstrates that condition overlap across different categories (e.g., conditions grouped in different dimensions), naturally forming distinct sub-groupings of states distinct from the original 14 dimensions.

**Figure 20**. Modularity community detection. Two distinct communities were identified, with 14 unique cliques and 18 total in the conditions network. Clique membership represents the total additional cliques that condition is also a member. A clique represents a unique subnetwork nested within a network.

The cross-consistency assessment and interdependence evaluation highlights that many dimensions and conditions are strongly correlated and linked to the others by varying degrees. Indeed, a collection of conditions can have strong positive or negative correlations, depending on the combination. Thus, it is reasonable to assess that rapid advancement in one dependent condition—such as the development of a new AI paradigm—can influence technical safety practices, race dynamics, and overall risk. At the same time, as evidenced by the high impact and likelihood values, tightly webbed across community sub-networks, transition velocity and diffusion are key state changes to evaluate system development and future capabilities.

Having developed the key dimensions and analyzed the consistency and interdependence, the averaged scenario pair assessment values are used to develop like-groupings of similar scored and correlated values. This study averages the assessment scores from one condition with every other to arrive at novel scenario combinations with differential interdependence, likelihood, and impact. For example, the *AGI capability* (0.74 impact and 0.51 likelihood) and *competitive takeoff* condition (0.5 impact and 0.55 likelihood) result in a combined scenario-pair impact value of 0.6 and a likelihood value of 0.53 for one unique configuration. Next, a clustering algorithm is used to combine the groupings into distinct scenario combinations. These values are then grouped into possible futures. Figure 21 shows the scenario pair combination process, with the condition's combined values at the intersection of each cell.

| | | | 1 | 2 | 3 | 4 | 5 | 6 | 7 | 8 | 9 | 10 | 11 | 12 | 13 | 14 | 15 | 16 | 17 |
|---|---|---|---|---|---|---|---|---|---|---|---|---|---|---|---|---|---|---|---|
| | | | Capability | | | | Diffusion | | | Transition | | | | Accelerant | | | Timeline | | |
| | | | low | Moderate | AI Ecology | AGI | Decentralized | Multipolar | Centralized | Slow | Moderate | Competitive | Fast | Overhang | Insight | Embodiment | Under 20 | 20-40 | Over 40 |
| | Dimensions | Conditions | | | | | | | | | | | | | | | | | |
| 1 | Capability | Low | | | | | | | | | | | | | | | | | |
| 2 | | Moderate | | | | | | | | | | | | | | | | | |
| 3 | | AI Ecology | | | | | | | | | | | | | | | | | |
| 4 | | AGI | | | | | | | | | | | | | | | | | |
| 5 | Diffusion | Decentralized | 0.37 | 0.55 | 0.57 | 0.50 | | | | | | | | | | | | | |
| 6 | | Multipolar | 0.45 | 0.63 | 0.65 | 0.58 | | | | | | | | | | | | | |
| 7 | | Centralized | 0.32 | 0.50 | 0.52 | 0.45 | | | | | | | | | | | | | |
| 8 | Transition | Slow | 0.29 | 0.47 | 0.49 | 0.43 | 0.40 | 0.49 | 0.36 | | | | | | | | | | |
| 9 | | Moderate | 0.39 | 0.57 | 0.59 | 0.53 | 0.51 | 0.59 | 0.46 | | | | | | | | | | |
| 10 | | Competitive | 0.40 | 0.58 | 0.60 | 0.53 | 0.51 | 0.59 | 0.46 | | | | | | | | | | |
| 11 | | Fast | 0.34 | 0.52 | 0.54 | 0.47 | 0.45 | 0.54 | 0.40 | | | | | | | | | | |
| 12 | Accelerant | Overhang | 0.41 | 0.59 | 0.61 | 0.54 | 0.52 | 0.60 | 0.47 | 0.44 | 0.55 | 0.55 | 0.49 | | | | | | |
| 13 | | Insight | 0.47 | 0.66 | 0.68 | 0.61 | 0.58 | 0.66 | 0.53 | 0.51 | 0.53 | 0.53 | 0.47 | | | | | | |
| 14 | | Embodiment | 0.39 | 0.57 | 0.59 | 0.52 | 0.50 | 0.58 | 0.45 | 0.42 | 0.53 | 0.53 | 0.47 | | | | | | |
| 15 | Timeline | Under 20 | 0.47 | 0.58 | 0.62 | 0.62 | 0.45 | 0.53 | 0.40 | 0.38 | 0.48 | 0.48 | 0.42 | 0.49 | 0.56 | 0.47 | | | |
| 16 | | 20-40 | 0.48 | 0.59 | 0.55 | 0.63 | 0.56 | 0.64 | 0.51 | 0.49 | 0.59 | 0.59 | 0.53 | 0.60 | 0.67 | 0.58 | | | |
| 17 | | Over 40 | 0.36 | 0.47 | 0.43 | 0.50 | 0.46 | 0.54 | 0.41 | 0.38 | 0.48 | 0.48 | 0.43 | 0.49 | 0.56 | 0.48 | | | |

**Figure 21**. The scenario pair combinatorial process. Each cell in the example matrix is a calculation between four values: 1) an impact and likelihood value for condition one (e.g., decentralized), and 2) an impact and likelihood value for condition 2 (e.g., AGI). The combined value for the conditions decentralized and AGI is 0.5 (top right in the first column). The values are then averaged and clustered for scenario development.

The GMA clustering process can result in millions of possible scenario combinations, or scenario pairs. After controlling for redundancy and inconsistent relationships, the final scenario pairs used in the model total 1,120. These pairs are then clustered using the Python Sci-kit learn library of clustering algorithms (Pedregosa, et al., 2011). The best choice for clustering like values without categories or data labels is unsupervised machine learning, as the model works to identify, from the bottom up, patterns in the data to group values that statistically appear most similar. There are multiple clustering algorithms available on Sci-kit learn. However, for this data type, the best choice is the k-means algorithm (Pedregosa, et al., 2011). The k-means algorithm is a good choice in that it easily uncovers subtle patterns in noisy data with manageable training sets (Dobilis, 2021).

## 5. Inferred Possible Futures

From the scenario development, value-pair combination, and clustering process, this study arrives at four scenario options that span the universe of plausible futures. The clustering process separated condition values from the lowest score to the highest into four combination groupings (Figure 22). At this stage, the clusters are evaluated to align the highest score for each condition to one of the four groupings. The futures described below synthesize a range of AI scenario variations. Risks to international security are examined across the spectrum from misuse to decision erosion and loss of control to distributed power-seeking and superintelligence. The four completed scenario combinations include: 1) a slow decentralized takeoff scenario, 2) an accelerated, uncontrolled, moderate transition with multipolar diffusion, and 3) a widely distributed unexpected takeoff (e.g., adaptive AI ecosystem), and 4) a fast centralized takeoff scenario (e.g., AGI agent). While some futures are easy to extrapolate, others are less intuitive; however, all scenarios can be envisioned along a spectrum from possible to plausible to probable.

| Alternative futures | Conditions Assigned to Each Scenario - from Low to High Impact & Likelihood | | | | |
|---|---|---|---|---|---|
| | Conditions – Moderate Impact & Low Likelihood | | | | |
| *Balancing Act* | Low Capability | Decentralized | Safety decrease | Isolation | Goal alignment |
| | Slow takeoff | New Paradigm | Over 40 years | Misuse | Individual |
| | Other | Current approach | Weak | Insight | Internet |
| | Conditions - Low Impact & Moderate Likelihood | | | | |
| *Accelerate Change* | Moderate power | Multipolar | Moderate takeoff | Hybrid Paradigm | Embodiment |
| | Over 40 years | Cooperation | Structural | Inner alignment | Coalition |
| | US-EU | Scale invariant | Strong governance | Safety ideal | Networks |
| | Conditions – Low Impact & Moderate Likelihood | | | | |
| *Shadow Intelligent Networks* | AI Ecology | Multipolar | Competitive | Hybrid paradigm | Insight |
| | 20-40 years | Failure | Goal alignment | Nation | US-EU |
| | General approach | Strong governance | Safety decrease | Monopolization | Networks |
| | Conditions – High Impact & Moderate Likelihood | | | | |
| *Emergence* | Under 20 years | Custom approach | Fast transition | Current paradigm | Institution |
| | AGI | Centralized | Failure | Overhang | Asia-Pacific |
| | Current paradigm | AI Arms race | Weak governance | Influence seeking | Narrow convergence |

**Figure 22**. Final scenario score card that aligns the given condition to each future. The scenarios have their unique risks but overall move from lowest impact – *Balancing Act* – to highest impact and surprise with *Emergence*.

## 5.1 Scenario A: Balancing Act - Low Impact/Moderate Likelihood

The future Balancing Act considers a low capability scenario with decentralized diffusion centered regionally in the US or EU (Figure 20). Following the COVID-19 pandemic, there is a retreat from globalization and a return to protectionism and isolation. As nations turn inward, major AI technology companies lose their global dominance in innovation. Increased international tensions and the decline in cooperation trigger sporadic outbreaks of conflict, with increasingly complex cyber-attacks being the primary risk. Nations must manage a delicate balance between cyber offense and defense as AI is increasing leveraged towards misuse. An individual or group of innovators discover a path to a new AI paradigm with the potential for high-level capabilities but with a long development time horizon. This scenario is preferable relative to the other potential futures, with a low likelihood of imminent high-level AI development. This alternative future expects the arrival of advanced AI in over 40 years, after approximately 2062. As nations turn inward and race dynamics stabilize, online knowledge communities explore new AI paradigms arriving at potential solutions through quantum machine learning. The scenario scores are medium to low on likelihood, considering the race dynamics of isolation, incremental timeline, and low capabilities. With low to moderate impact scores, this scenario fits into a preferred category with respect to overall disruptive capacity.

| Balancing Act: Moderate Impact/Low Likelihood ||||||
| Dimension | Condition | Status | Impact | Likelihood |
| --- | --- | --- | --- | --- |
| Capability | Low | The corporate AI sector plateaus. Research continues but toward new ideas. | 0.41 | 0.27 |
| Diffusion | Decentralized | Computational resources are widely distributed, leveling the field to exploratory innovation. | 0.4 | 0.49 |
| Transition | Incremental | The pause in corporate dominance slows the pace of investment and development. | 0.2 | 0.33 |
| Paradigm | New paradigm | Deep learning plateaus, initiating research in alternative AI models, such as quantum ML. | 0.34 | 0.6 |
| Accelerant | Insight | Through incentivized innovation, scientists discover a means to stabilize quantum computation. | 0.3 | 0.71 |
| Timeline | Over 40 years | The pace of competitive development slows, with exploratory basic research prominent. | 0.34 | 0.43 |
| Race dynamics | Isolation | Global markets take an inward turn. Big technology companies decrease resource-intensive AI development. | 0.6 | 0.49 |
| Risk | Misuse | Tensions are at an all-time high internationally and AI has provided new means for cyber attack strategies. | 0.71 | 0.62 |
| Technical risk | Inner alignment | While cyber remains the dominant threat, inner alignment failures continue to plague AI systems. | 0.56 | 0.57 |
| AI safety | Scale invariant | To manage alignment risks, iterated amplification and IRL prove effective for goal alignment. | 0.76 | 0.53 |
| Developer | Individual | With a wider distribution of resources, greater opportunities are available for individuals. | 0.56 | 0.31 |
| Region | Other | With lower resource requirements, researchers globally enjoy broad opportunities. | 0.53 | 0.27 |
| Governance | Strong | Corporate decline is accompanied by multilateral governance and international regulations. | 0.73 | 0.44 |
| Corporate governance | Safety decrease | Common safety standards and coordination slump with the decentralized market. | 0.79 | 0.39 |
| Average | | | 0.535 | 0.474 |

Figure 23. Scenario-A overview. Balancing Act portrays a future where globalization and competitive AI development decline in prominence. As companies decrease basic research, networks of scientists explore alternate paradigms for AI development.

## 5.2 Scenario B: Accelerating Change

Accelerating Change presents one of the more plausible and middle-of-the-road alternative futures, given our current understanding (Figure 24). Several of the conditions in the scenario broadly align with current AI research and are relatively easy to extrapolate. Centered in the West, Accelerating Change envisions a moderate competitive transition, with multipolar diffusion, as technology giants consolidate control over various industries. Increasingly powerful AI monopolies lead to substantive shifts in power dynamics, from the nation state to the corporation as the new dominant power structure. This future expects the arrival of advanced AI in 20 and 40 years, between approximately 2042 and 2062. Participants ranked the likelihood values for moderate capability, competitive takeoff, and multipolar diffusion at or above 50 percent. Western origin, moderate power, monopoly control, and structural risks all ranked above 60 percent. This future explores the structural risks to society from technologies that accelerate faster than institutions or individuals can collectively manage. Several conditions in the scenario (negatively correlated with the dominant categories) are less intuitive and are infrequently included in similar scenarios, allowing unique combinations. For example, embodiment is the key driver of the AI transition, and safety standards and governance are in positive territory despite the geopolitical environment.

| Dimension | Condition | Status | Impact | Likelihood |
|---|---|---|---|---|
| **Accelerating Change: Low-Impact & Moderate Likelihood** | | | | |
| Capability | Moderate | AI research maintains momentum and accelerates with systems capable of moderate generalization. | 0.47 | 0.62 |
| Diffusion | Multipolar | The latest breakthroughs in general AI are available for leading companies, allowing companies to maintain parity. | 0.4 | 0.49 |
| Transition | Moderate uncontrolled | Capability jumps increase bringing novel innovations faster than society can keep pace. | 0.6 | 0.55 |
| Paradigm | Hybrid | Researchers in several coalitions discover the means to initiate machine common sense. | 0.42 | 0.72 |
| Accelerant | Embodiment | Leaders in two coalitions discover a means for virtual and physical embodiment to stimulate generalization. | 0.47 | 0.54 |
| Timeline | 20-40 Years | The pace of development increases rapidly, but steadily, posed to reach the human level in the 2040 timeframe. | 0.59 | 0.66 |
| Race Dynamics | Cooperation | Large AI corporations cooperate on the latest research to assist in reaching milestones. | 0.13 | 0.37 |
| Risk | Structural | The advancing wave of change brings difficulties in societal stability, decision making, and psychology. | 0.58 | 0.61 |
| Technical Risk | Inner alignment | While cyber remains the dominant threat, inner alignment failures continue to plague AI systems. | 0.56 | 0.57 |
| AI Safety | Scale invariant | To manage alignment risks, iterated amplification and IRL prove effective for goal alignment. | 0.34 | 0.42 |
| Developer | Coalition | Large technology companies, wielding more power than most states, create agglomerations of corporate alliances. | 0.15 | 0.38 |
| Region | USA-EU | Corporate leaders in the West and South Asia maintain the lead in AI innovation. | 0.2 | 0.78 |
| Governance | Strong | While geopolitics remain fraught, AI governance is on track and managing proper use between nations. | 0.15 | 0.43 |
| Corporate governance | Safety Ideal | Safety standards and coordination keep pace with the changing technology. | 0.12 | 0.49 |
| Average | | | **0.37** | **0.545** |

**Figure 24**. Scenario-B Overview. Accelerated Change depicts a future where multipolar blocks of nations compete for AI dominance, while technology accelerates faster than institutions can manage. AI systems can generalize moderately but bring transformative capabilities.

*5.3 Scenario C: Shadow Intelligent Networks*

The Intelligent Networks' future is one of the more subtle scenarios and was ranked between somewhat and very likely by participants but highly variable across conditions (Figure 25). This scenario class (like Accelerating Change) lines up well with our current future trajectory in 2022. This scenario describes a low-capability, widely distributed network of intelligent agents, and is closely modeled on Eric Drexler's "Comprehensive AI Services" (CAIS) framework and the AI ecology class of scenarios (Drexler, 2019). The focus is concentrated on potential risks from self-organizing adaptive agent communities and issues of control, considering planned smart grid and next-generation networks technologies (GCN Staff, 2021) (Radanliev, De Roure, Van Kleek, Santos, & Ani , 2021). The AI ecology condition was evaluated at a higher-than-expected likelihood of 66 percent, the highest score for that dimension. The future imagines an AI ecology emergence and distributed diffusion using a hybrid AI paradigm. The possible triggers include an insight into intelligence, with the primary risk being goal alignment and the potential risks from runaway distributed systems. Like the previous future, monopolization and multipolar diffusion are key aspects of the emergence, with the West being the region of origin. Goal alignment, hybrid paradigm, insight, and general AI safety techniques all scored above 70 percent for likelihood. Experts judged the arrival of advanced AI in this scenario in 20 to 40 years, between approximately 2042 and 2062.

| Shadow Intelligent Networks: Low-Impact & Moderate Likelihood | | | | |
|---|---|---|---|---|
| Dimension | Condition | Status | Impact | Likelihood |
| Capability | AI ecology | AI is deployed as a decentralized system of agents and services that can generalize and recursively improve collectively. | 0.32 | 0.66 |
| Diffusion | Multipolar | The latest breakthroughs in general AI are available for leading companies, allowing companies to maintain parity. | 0.56 | 0.66 |
| Transition | Competitive | AI services model is competitively sought to gain an advantage in a moderate, controlled rapid takeoff. | 0.5 | 0.55 |
| Paradigm | New paradigm | Researchers discover a means to generalize across an ecosystem of individual agents and services. | 0.34 | 0.6 |
| Accelerant | Insight | Through distributing thousand of agents across intelligent next-generation networks, AI systems show capability gains. | 0.3 | 0.71 |
| Timeline | 20-40 years | The pace of development increases rapidly, assessing approximate gains in general intelligence in 20 years. | 0.59 | 0.66 |
| Race Dynamics | Monopolization | Large AI corporations cooperate on the latest research to assist in reaching milestones. | 0.71 | 0.63 |
| Risk | Failure | The widely distributed intelligence brings occasional disturbances and intermittent failures. | 0.7 | 0.74 |
| Technical Risk | Goal alignment | With the large number of competing agents, transmitting clear objectives without interference is problematic | 0.56 | 0.57 |
| Safety Technique | General approach | To align agents, AI safety researchers must revisit first principles to align the instantiations. | 0.55 | 0.69 |
| Developer | Nation | Researchers discover a means to generalize across an ecosystem of individual agents and services. | 0.39 | 0.6 |
| Region | US-EU | Western nations develop the means for the first intelligent networks with autonomous adaptive agents. | 0.2 | 0.78 |
| Governance | Strong | Within allied countries, AI governance is strong with limited international governing bodies and regulations. | 0.34 | 0.49 |
| Corporate Governance | Safety decrease | Safety standards and coordination have increased in individual companies but limited safety coordination. | 0.29 | 0.54 |
| Average | | | 0.463 | 0.634 |

Figure 25. Scenario-C overview. Shadow Intelligent Networks describes a future where researchers discover that generalization is possible through machine-to-machine communication across intelligent networks. This future is considered low impact and moderate likelihood by researchers.

*5.4 Scenario D: Emergence*

The alternative futures Emergence is assessed less likely than the other three scenarios but with greater consequences (Figure 26). There are variations between conditions, with respect to plausibility and impact, but cumulatively the future is scored as somewhat unlikely, but with very high impact. Emergence envisions a fast takeoff scenario in a non-Western state, in an economic and geopolitical AI arms race environment. Advanced AI capabilities are resource prohibitive, centralized, and compartmentalized within major companies or government special programs. AI technologies and intellectual property have been labeled a national strategic asset, stoking the arms race dynamic, and as research develops some major companies either work with the government and military or are nationalized. High-level systems are expected within a 20-year window before 2042. The primary safety concerns are system failures, inner misalignment, and power-seeking. Fast takeoff, centralized diffusion, and an abrupt 20-year timeframe led by regional powers in East or South Asia scored at or below 50 percent probability. However, each scored above 60 for impact, with fast takeoff and centralized diffusion categorized very high. The technical risk for this future power-seeking is one of the highest scored values for impact and likelihood.

| \<td colspan=5\> Emergence: High-Impact & Moderate Likelihood | | | | |
|---|---|---|---|---|
| **Dimension** | **Condition** | **Status** | **Impact** | **Likelihood** |
| Capability | AGI | Rapid scale-up in compute leads to new capabilities to generalize. Compute overhang produces the first prosaic AGI. | 0.74 | 0.51 |
| Diffusion | Centralized | The latest breakthroughs are controlled by research groups of large companies or covert government projects. | 0.75 | 0.38 |
| Transition | Fast takeoff | Compute overhang unleashes radical discontinuity in the system's cognitive ability in a short timeframe. | 0.88 | 0.45 |
| Paradigm | Current paradigm | Researchers learn that scaling laws are enough to bring novel capabilities to AI systems. | 0.76 | 0.5 |
| Accelerant | Overhang | An increase in scale brings new capabilities periodically until a critical threshold, where cognitive capabilities jump from 1-100. | 0.71 | 0.56 |
| Timeline | Under 20 years | The pace of development increases steadily until compute overhang is discovered. | 0.78 | 0.42 |
| Race Dynamics | Arms Race | Large AI corporations are leveraged by the government and military to ensure technological dominance. | 0.86 | 0.67 |
| Risk | Misuse | Governments use proxy cyber threat actors to execute attacks against adversaries with new AI tools. | 0.71 | 0.62 |
| Technical Risk | Influence | As AI systems increase in strength they begin to exhibit influence-seeking behavior. | 0.84 | 0.65 |
| AI Safety | Custom approach | For the newly powerful systems, custom techniques must be developed for each unique instantiation. | 0.76 | 0.53 |
| Developer | Institution | Leading companies control AI assets but act as an arm of the government and military. | 0.77 | 0.7 |
| Region | Asia-Pacific | China takes the lead in AI development in partnership with other nations in their belt and road area of influence. | 0.66 | 0.54 |
| Governance | Weak | As competition heats up into an all-out war for AI leadership AI governance weakens. | 0.73 | 0.44 |
| Corporate Governance | Safety Decrease | Safety standards take a back seat as competition becomes the primary goal. | 0.79 | 0.39 |
| Average | | | 0.767 | 0.525 |

**Figure 26.** Scenario-D overview. The alternative future Emergence describes a future where the current AI paradigm continues making consistent capability jumps and at a certain critical point a compute overhang initiates a radical gain in cognitive capacity.

# 6. Discussion: The Risk Landscape

The range of potential AI risks is broad, from social instability and value erosion to unexpected accidents, cascading failures, and collapse. There is no shortage of dangers. Many of the more extreme risks are downplayed as unconvincing, highly improbable, or impossible by some researchers, but as systems continue to scale to new milestones, more concerted attention is warranted. While some of the AI scenarios are highly speculative, they are grounded in current research and within the range of possible futures. Indeed, if there is a nonzero chance of an extreme AI scenario—unintentionally escalating conflict, shifting the balance of power, or compromising control—leaders must challenge assumptions, incorporate uncertainty, and test the boundaries of what is possible.

AI system unpredictability, manifesting agential characteristics with goal-directed behavior, is not strictly conceptual but has been observed experimentally. Examples include AI game-playing agents that learned to alter their environments and resources (e.g., weapons for game-playing AI), find loopholes in the system or reward function, and develop dangerous or unexpected subgoals as optimal strategies to complete a task. Indeed, AI systems learning the wrong goal or developing instrumental subgoals are key concerns of AI safety research. It is not that an agent will consciously strive to harm, but rather it will complete a task in a way that indirectly causes harm. This could be as simple as subverting the rules of a game, or as severe as destabilizing a power grid or nuclear weapons system. Being able to determine if an objective is learned or specified incorrectly is difficult, often requiring that the problem manifest first before it can be addressed. These problems will be highly dangerous as AI is further integrated into safety-critical systems, military infrastructure, or strategic weapons.

This could be especially problematic with inner alignment (or mesa optimization), where the AI is optimized for a specific objective, but the model learns a proxy or approximate objective, hidden in the neural network. Examined closely, inner alignment appears hopeless. While outer alignment concerns the misalignment of a defined objective (the base objective), inner alignment is where the learned model is itself optimizing for a separate objective (the mesa objective), undetectable to the programmer until the behavior manifests. A system could be determined trustworthy by developers yet fail unexpectedly due to a minor change in deployment. For example, if an autonomous weapons system is performing well in training and testing but fails in deployment, it is too late. We must understand if a system will interpret instructions correctly—or pursue a proxy or invalid objective instead—and how to know the difference before a catastrophe occurs.

Influence or power-seeking appears especially concerning, ranked 0.84 for impact and 0.65 for likelihood by domain experts in the survey. Research published by OpenAI in 2019 demonstrated this behavior in a simulated environment: two teams of agents, instructed to play hide-and-seek, proceeded to develop independent strategies, coevolve tactics, and horde objects from the competing team, in what OpenAI described as "emergent tool use" (Baker, et al., 2020). The agent's tactics increased in sophistication through each iteration, constructing shelters and using ramps for offense and defense. This research is a good example of systems learning—independently, without explicit instructions—to collect resources, develop instrumental subgoals, and strategize to win. This should give pause to leaders seeking to integrate AI systems into critical networks. In 2008, computer science professor Steve Omohundro illustrated the innate drives that systems will pursue "unless explicitly counteracted," providing the theoretical framework for understanding risks from artificial agents (Omohundro, 2008) (Bostrom, 2014). Since that time, research has shown that these are indeed tangible concerns.

Collecting resources to strategize in a game is not a significant risk, but this depends on the complexity of the objectives. For example, consider an AI system in charge of managing power to and from two towns. Without specific programming to the contrary, the agent could hack into external computational resources for efficiency or redirect power from a third town to optimize the objective (Carlsmith, 2022). Having materialized similar behavior at a rudimentary level, it is uncertain whether more capable agentic systems would not manifest similar traits in more complex circumstances. With the increased use of multiagent environments—and proposals for multiagent self-organizing networks—an evaluation of the game-theoretic risks of this paradigm are worth consideration (Radanliev, De Roure, Nicolescu, Huth, & Santos, 2022) (Berggren, et al., 2021) (Nguyen & Reddi, 2021).

The unpredictability of AI with unique world models will inevitably introduce surprises: novel solutions that benefit humanity but also unexpected ones with disruptive consequences. Besides AI system misspecification or misalignment, there are modern precedents of system failures that outline the degree of unpredictability. For example, consider the flash crash on wall street in 2010 that drove the market down 600 points in 5 minutes, or the UK in 2016, where the pound sterling dropped 8 percent of its value in minutes (Townsend, 2016). This example, and many others, point to the problems with algorithmic control. Over time, we normalize successful algorithms and do not question their accuracy or decision processes if demonstrated as reliable. However, this becomes dangerous as AI is weaved deeper into industry and government. The days of the busy trading floor at the New York Stock Exchange are over. High-frequency trading is the norm, where trades take a fraction of a second and are initiated from subtle patterns in the data, too small for a human to perceive (Scharre, 2019, p. 189). Thus, based on the objective defined by the trader, the algorithm optimizes the best strategy, which as described can take shape in any number of ways with a proxy or approximate solution, or could be misunderstood entirely. Once AI systems are fully integrated into the decision environment, deconfliction becomes extremely challenging.

The structural risks examined in the scenario "Accelerated Change" are visible today. At first glance, structural risks seem less of a direct threat to international security, but they are prominent forces that will increase instability (much like climate change, disease, and migration) and have secondary and tertiary effects down the line. For example, consider social media. The impact of algorithmic interactions on relationships and decision-making is a key force that will influence purchase choices, social interactions, and politics. Indeed, there is evidence (at least in part) that people's interests, decisions, and beliefs are increasingly driven by algorithm recommendations (what is termed "nudging"). Incremental but consistent nudging can result in big changes over time. Thus, the erosion of human decisiveness and alteration of preferences is a risk that could span generations.

# 7. Conclusion

Progress in AI is advancing faster than society, institutions, and many researchers can keep up. The pace of AI milestones, discovery of new scaling laws, and potential avenues for advanced generalization are accelerating at an exciting but somewhat unnerving rate. Warnings of an imminent AI winter and the decline of Moore's law continue, and these proclamations could be accurate; however, the virtuous circle of feedback loops, complementary technologies, financial investments, talent, and self-improving AI systems continues. Thus, evaluating the range of plausible futures sooner rather than later is critical so that leaders can plan accordingly. At the same time, maintaining visibility of the movement of AI conditions, their interactions and directionality, could help analysts keep track the overarching trends of the technology. While forecasting specific trajectories is untenable, understanding the broad outlines and potential sharp left turns could help ensure societal and institutional resilience.

The expert elicitation found some unexpected insights and others that reinforced the overall body of literature. For example, the probability of influence-seeking was ranked far higher than expected, especially given the potential impact. Power-seeking consistently scored above 65 percent likelihood, while ranking close to 80 for overall impact on safety and security. The high likelihood speaks to the increasing evidence of this emergent behavior in AI systems, as noted above with OpenAI's tool use discovery. Similarly, goal alignment and overall failure modes were ranked at 74 percent likely and 70 for impact. While not a complete surprise, given the growth of the field and experimental data, the high numbers do highlight the seriousness of the risks. Indeed, while it is unpopular to blame the system, it is a basic truth that AI systems are operating through parallel cognitive architectures, producing solutions far removed from human understanding. The latent space of AI decision processes produces highly capable results but are ultimately opaque. The more AI systems are charged with making impactful decisions, whether negotiating insurance prices or a criminal conviction, the more we must admit we're not fully in control. These technical problems are unlikely to be resolved by governance measures or regulations, and likely requiring technical solutions. The other high likelihood and impact condition was the potential for an AI arms race.

In addition, there is a strong belief that the big technology corporations will remain the preeminent institutions to discover advanced AI and that the trend toward monopolization and industry control will continue. A noticeable trend that varies from previous surveys in the AI risk community is the decline in the belief that advanced AI will be discovered and controlled by a solitary institution or research group, or that a fast takeoff will be the most probable means of transition. This change can similarly be observed in the literature over the past three years (Sotala, 2018). The participant responses suggest advanced AI is trending toward a moderate capability in the near term, involving a variety of actors, with a rapid build, or "surge" transition as described by Max Moore, rather than an abrupt discontinuity measured by hours (More, 2009). AI safety was a very niche area of research for many years, and remains so comparatively, but has experienced explosive growth in recent years as AI has been increasingly interlaced through society, institutions, and government. At the same time, ethical concerns, ubiquitous surveillance, and substantive risks from failure modes have presented in kind, driving many researchers to pursue means to curtail potential dangers.[4] As capabilities continue to progress, there is much uncertainty as to the pace of change and at what cost.

---

[4] Researchers at some of the top global AI companies have taken the lead to align AI with human values, including OpenAI (https://openai.com/alignment/), Deep Mind (https://www.deepmind.com/publications/artificial-intelligence-values-and-alignment), and a number of top universities in the forefront of AI research including Berkely's Center for Human Compatible Artificial Intelligence (CHAI, https://humancompatible.ai/) and Stanford University's Human-Centered Artificial Intelligence (HAI, https://hai.stanford.edu/).

As we move toward this uncertain future, the risk of failures, goal misspecification, misalignment, or malicious use by a state or non-state group is unnervingly high but variable across differential technological paths. The combination of each variation paints a complex picture. The benefits of advanced AI will be a game changer for prosperity if managed safely or pose grave unpredictable dangers if not. These risks are recognized and are increasingly being evaluated by researchers. A large body of work in the national security enterprise is appropriately focusing on lethal autonomous weapons (LAWS), but the focus has been almost exclusively on issues of trust, misuse, or accidents, (Flournoy, Haines, & Chefitz, 2020) with notable exceptions by AI risk researchers and institutes (Baum & Barrett, 2017) (Geist & Lohn, 2018) (Ding & Dafoe, 2021) (Shahar & Amadae, 2019). The focus must expand beyond ethics, misuse, and accidents and consider risks from the systems themselves and the agential dangers inherent in goal-directed systems. Like many complex systems, AI tends to spread beyond its original intent, and humanity must be vigilant in understanding and mitigating the risks that this presents.

## 8. Acknowledgements

The authors would like to acknowledge Dr. David Blauvelt, Dr. Stacy Closson, Dr. Adam Jungdahl, and Dr. John Greer for their insights and helpful discussions. This research did not receive any specific grant from funding agencies in the public, commercial, or not-for-profit sectors. The authors have no conflicts of interest to report.

# Appendix

*Survey Questions*

1. Please provide your best assessment of the likelihood of the following condition occurring.
2. If the following condition were to occur, what overall impact would you expect it to have on society and security?

## Dimension And Condition Definitions

*Capability and generality*

1. **Low**: AI Systems remain approximately as capable and general as current systems and progress only marginally in power and general-purpose capabilities. Decreased investment and an AI winter are possible
2. **Moderate**: AI systems become increasingly powerful and generalizable across multiple cognitive tasks in a range of fields. Society and institutions struggle to keep pace with the rate of change and complex optimization processes.
3. **AI Ecology:** Systems develop rapidly across a digital ecosystem of interacting AI systems and agents. Systems adapt and evolve to human and superhuman capable across multiple domains (a variation of the Comprehensive AI services model - CAIS)
4. **AGI:** AI systems progress to an approximate human-level AGI. The system is as capable and general as humans in all domains. With computational and memory advantages, the AGI is capable of recursive self-improvement to ASI.

*Transition*

1. **Slow**: AI systems develop incrementally and there is the possibility of an AI winter. Powerful capabilities are theoretically possible, but they develop over a much longer time horizon (decades or longer).
2. **Moderate uncontrolled** (continuous): Systems develop rapidly but with no sharp discontinuity. The changes spread faster than anticipated with surprising capability jumps that are extremely difficult for society to manage or understand (months or years to less than a decade).
3. **Moderate competitive** (continuous): Systems develop rapidly (no sharp discontinuity). Radical changes are anticipated and actively pursued, for competitive advantage, including the lead-up or response to conflict. Unexpected capability jumps but control efforts are planned. This scenario is related to highly competitive race dynamics and could have geopolitical dimensions (many months to years, less than a decade).
4. **Fast** (discontinuous): System(s) develop rapidly and at an approximate human-level capability and generality undergoes a radical shift in power from AGI to artificial superintelligence (ASI) through recursive self-improvement (minutes, hours, days).

*Diffusion*

1. **Decentralized**: AI systems are widely available through open-source networks when HLMI is developed. Resource requirements are low, bringing inordinate power to citizens.
2. **Multipolar**: AI discoveries are made across leading companies only, with technological parity and resources, in several countries. Multipolar scenario.
3. **Centralized**: The system is discovered by and confined to one lab or government program. This includes scenarios where the discovery is part of a corporation's special program (e.g., Google X), a surprise discovery, or an accident.

*Timeframe*

1. **Less than 20 years:** High-level machine intelligence or a close approximation is developed before 2040. The system is capable to complete most cognitive tasks of a human being. This includes the possibility of AGI or ASI but does not depend on that exact instantiation.
2. **20 to 40 years:** High-level machine intelligence or a close approximation is developed sometime between 2035 and 2070 The system is capable to complete most cognitive tasks of a human being. This includes the possibility of AGI or ASI but does not depend on that exact instantiation.
3. **Greater than 40 years:** High-level machine intelligence or a close approximation will take over 40 years to develop. The system is capable to complete most cognitive tasks of a human being. This includes the possibility of AGI or ASI but does not depend on that exact instantiation.

*Accelerants*

1. **Compute overhang:** A new algorithm, overlooked insight, or paradigm exploits existing computational resources far more efficiently than previously, allowing rapid gains in capability or generality.
2. **Innovation:** A new insight, machine learning paradigm, or completely new architecture accelerates capabilities, from 0 to 100, allowing must faster and more general capabilities. Examples could include insight from neuroscience, a new mode of learning (e.g., common sense), or quantum materials or computation.
3. **Embodiment/Data:** Simulated or actual embodiment, a new type or quality of data for ML training provides radical capability gains.

*Paradigm*

1. **Current paradigm:** The current machine learning paradigms can scale up radically to advanced capabilities with broad generality, up to and including AGI ("prosaic AGI")
2. **New approach:** High-level systems requires an entirely new AI paradigm. New modes of learning such as system two reasoning, a fundamental insight on intelligence, or new architectures are required to reach high-level general decision making.
3. **Hybrid approach:** Advanced general AI systems are attainable using current machine learning paradigms but require something else. Current learning methods are on the right track but require additional learning techniques, such as a hybrid approach, common sense reasoning, genetic algorithms plus self-supervised learning.

*Race Dynamics*

1. **Cooperation:** AI technologies are recognized as a global public good and cooperation increases between companies and national governments. Race to the top scenario.
2. **Isolation:** Global governments take a protectionist turn and cooperation decreases. AI is developed in isolation. Markets attempt to maintain the status quo and companies compete regionally or within national borders, causing wide disparities in technical standards and regulations.
3. **Monopolization:** Technology companies increase acquisitions of smaller companies and talent to control AI resources. Corporations increasingly control the direction of research, influence over governments, and distribution of power. In the extreme, companies become semi-sovereign entities beyond the reach of government and international institutions.
4. **AI Arms Race:** AI is named a strategic national asset and countries race for global dominance. As high-level capabilities become more likely, governments begin to control research and access and use top companies as an arm of military power. AI is militarized and conflict is more likely.

*Dominant Risk Class*

1. **Misuse:** Alignment is under control and Cyber-attacks and disinformation campaigns increase in frequency and disruptive potential. Persistent surveillance becomes more likely by governments and criminals.
2. **Failures:** AI systems are given more control over decision processes making failure modes more consequential and goal alignment remains the key danger. With systems in control of increasingly sensitive infrastructure, a failure could result in cascades of follow-on failures.
3. **Structural:** Increased decision autonomy of AI systems brings subtle changes to the functioning of society and uncertainty of conflict. Overlap between nations' offense/defense balance makes it more likely for military escalation. Values decline as AI takes control of all decision processes.

*AI Safety*

1. **Scale invariant**: Current AI safety techniques can scale to high-level systems. The current techniques being designed for modern ML are broadly transferable to high-level general systems.
2. **New approach**: New AI safety techniques must be developed from first principles to be effective against high-powered more general systems.
3. **Custom approach**: Each unique instantiation of an advanced AI system requires a specialized safety technique to be developed, making alignment a far more complex problem.

*Technical Safety Risk*

1. **Goal Alignment**: Goal alignment remains the primary intractable problem that we are unable to solve. Progress in alignment has had success, but system changes require entirely new solutions. The most dangerous risk from HLMI remains misaligned systems.
2. **Power-seeking:** The most prevalent and dangerous concern turns out to be the acquisition of resources by AI systems. Even with improvements to goal alignment, instrumental objectives, and deception to prevent changes, is difficult to detect and varied across all systems. The potential to lose control is high.
3. **Inner Alignment** (Mesa Optimization): Goal alignment has had significant success, but inner aligned agent models remain a problem and are extremely difficult to identify. Subtle and impossible-to-detect misalignment issues and failures remain prevalent and are the most dangerous concern.

*Actor*

1. **Coalition of states (e.g., EU, NATO):** A coalition of nation-states, international organizations, or military alliances develop the first radically capable advanced AI systems.
2. **Country:** An individual government discovers or develops radically transformative AI systems. This could be through a national government program, the military, or by nationalizing one or several corporations.
3. **Institution:** A private-sector corporation (e.g., Tencent, Google), non-profit, or academic research institution develops the first advanced AI instantiation.
4. **Individual:** A private developer discovers an advanced AI capability. This is more likely in circumstances where AI research and development remains open-source and resource requirements are low (e.g., a new AI paradigm).

*International Governance*

1. **Weak:** (decrease in governance) Preparation stays the same as today (reactive) or decreases in cooperation, collective action, and agreements due to isolationism or conflict and weakening of norms and institutions, possibly due to race dynamics.
2. **Moderate:** A strengthening of international norms and consolidation of institutions. International norms on the proper use of AI systems are well established and an agreed-upon framework of safety standards is established.
3. **Strong:** International safety regimes established (e.g., IAEA), multilateral agreements, and verification measures (e.g., IAEA nuclear inspections) enacted for states unwilling to sign on to AI safety agreements. An international body on AI safety is established that coordinates efforts.

*AI Safety Governance*

1. **Decrease**: An increase in economic competition brings decreased cooperation across leading AI companies, impacting safety coordination. Isolation could worsen this.

2. **Moderate**: AI companies and research institutions increase coordination on AI development and technical safety practices, with intercompany working groups on technical safety standards and control measures.
3. **Strengthen**: AI companies and research institutions agree on third-party safety standards and a common framework for technical safety control measures.

*Region*

1. **USA-Western European:** Major companies in the US or headquartered in the US or the EU develop the first HLMI instantiation. This region additionally includes close allies often considered "western" such as Australia and Japan.
2. **Asia-Pacific:** Greater Asia – South, Southeast, Southwest, and East – develop the first HLMI instantiation. This includes the pacific islands, Eurasia, Russia, and the Middle East.
3. **Africa or Latin America/Caribbean:** The global south, besides Asia. This includes Central, Sand outh America, the Caribbean, and continental Africa.

*Superintelligence Scenarios:*

1. **The internet as emergent intelligence:** Unable to recognize the qualitatively different forms of intelligence, the internet has been developing intelligence as a large complex system. The collective system sparks the emergence of a single superintelligence.
2. **Cognitive Internet-of-Things**: As AI is networked throughout all sensors and systems, machine agents proliferate across global networks as a sensor web of millions of independent agents, with independent alignment risks.
3. **Narrow AI systems convergence:** As tool AI continues to spread and increase in power (CAIS model), like strands of DNA, these individual agents combine and emerge as one superintelligence.

http://www2.mitre.org/work/sepo/toolkits/risk/ToolsTechniques/files/UserGuide220.pdf

Flournoy, M. A., Haines, A., & Chefitz, G. (2020, October 06). *Building Trust Through Testing: Adapting DOD's Test & Evaluation, Validation & Verification (TEVV) Enterprise for Machine Learning Systems, including Deep Learning Systems.* Retrieved from Georgetown University Center for Security and Emerging Technologies: https://cset.georgetown.edu/wp-content/uploads/Building-Trust-Through-Testing.pdf

GCN Staff. (2021, April 30). *NSF, NIST, DOD team up on resilient next-gen networking.* Retrieved May 1, 2022, from GCN: https://gcn.com/cybersecurity/2021/04/nsf-nist-dod-team-up-on-resilient-next-gen-networking/315337/

Geist, E., & Lohn, A. J. (2018). *How Might Artificial Intelligence Affect the Risk of Nuclear War?* doi:https://doi.org/10.7249/PE296

Golbeck, J. (2015). *Introduction to Social Media Investigation: A Hands-on Approach.* Waltham: Syngress.

Grace, K., Salvatier, J., Dafoe, A., Zhang, B., & Evans, O. (2018, July 31). Viewpoint: When Will AI Exceed Human Performance? Evidence from AI Experts. *Journal of Artificial Intelligence Research, 62*, 729-754. doi:https://doi.org/10.1613/jair.1.11222

Granger, M. M. (2014). Use (and abuse) of expert elicitation in support of decision making for public policy. *Proceedings of the National Academy of Sciences*, 7176-7184.

Grossman, G. (2022, June 04). *Is DeepMind's Gato the world's first AGI?* Retrieved from Venture Beat: https://venturebeat.com/datadecisionmakers/is-deepminds-gato-the-worlds-first-agi/

Gruetzemacher, R., & Paradice, D. (2019). Toward Mapping the Paths to AGI. *12th International Conference, AGI 2019* (pp. 70-79). Shenzhen: Springer International Publishing.

Hagberg, A. A., Schult, D. A., & Swart, P. J. (2008). Exploring network structure, dynamics, and function using NetworkX. In G. Varoquaux, T. Vaught, & J. Millman (Ed.), *Proceedings of the 7th Python in Science Conference (SciPy2008)*, (pp. 11-15). Pasadena. Retrieved from https://conference.scipy.org/proceedings/SciPy2008/paper_2/

Hawking, S., Russell, S., Tegmark, M., & Wilczek, F. (2014, May 01). *Stephen Hawking: 'Transcendence looks at the implications of artificial intelligence - but are we taking AI seriously enough?'.* Retrieved from The Independent: https://www.independent.co.uk/news/science/stephen-hawking-transcendence-looks-at-the-implications-of-artificial-intelligence-but-are-we-taking-ai-seriously-enough-9313474.html

Hernández-Orallo, J., Sheng Loe, B., Cheke, L., Martínez-Plumed, F., & ÓhÉigeartaigh, S. (2021). General intelligence disentangled via a generality metric for natural and artificial intelligence. *Nature Scientific Reports, 11*(22822). doi:https://doi.org/10.1038/s41598-021-01997-7

Omohundro, S. M. (2008). The Basic AI Drives. *Proceedings of the 2008 conference on Artificial General Intelligence 2008: Proceedings of the First AGI Conference* (pp. 483–492). Amsterdam: IOS Press.

Pedregosa, F., Varoquaux, G., Gramfort, A., Michel, V., Thirion, B., Grisel, O., . . . Duchesnay, É. (2011). Scikit-learn: Machine Learning in Python. *Journal of Machine Learning Research, 12*(85), 2825-2830. Retrieved from http://jmlr.org/papers/v12/pedregosa11a.html

Perry, L. (2020, June 15). *Steven Pinker and Stuart Russell on the Foundations, Benefits, and Possible Existential Threat of AI.* Retrieved from Future of Life Institute: https://futureoflife.org/2020/06/15/steven-pinker-and-stuart-russell-on-the-foundations-benefits-and-possible-existential-risk-of-ai/

Radanliev, P., De Roure, D., Nicolescu, R., Huth, M., & Santos, O. (2022). Digital twins: artificial intelligence and the IoT cyber-physical systems in Industry 4.0. *International Journal of Intelligent Robotics and Applications, 6*, 171–185. doi:https://doi.org/10.1007/s41315-021-00180-5

Radanliev, P., De Roure, D., Van Kleek, M., Santos, O., & Ani , U. (2021). Artificial intelligence in cyber physical systems. *AI & Society, 36*, 783–796. doi:https://doi.org/10.1007/s00146-020-01049-0

Reed, S., Zolna, K., Parisotto, E., Colmenarejo, S. G., Novikov, A., Barth-Maron, G., . . . de Freitas, N. (2022, May 19). *A Generalist Agent.* Retrieved from arXiv: https://arxiv.org/pdf/2205.06175.pdf

Ritchey, T. (2014). General Morphological Analysis * A general method for non-quantified modelling. Swedish Morphological Society. Retrieved from https://www.semanticscholar.org/paper/General-Morphological-Analysis-*-A-general-method-Ritchey/1c508d794dc86083cededa5cfddd144404a8d42e

Russell, S. (2019). *Human Compatible: Artificial Intelligence and the Problem of Control.* New York: Viking.

Scharre, P. (2019). *Army of None: Autonomous Weapons and the Future of War.* New York: Norton & Company.

Shahar, A., & Amadae, S. (2019). Autonomy and machine learning at the interface of nuclear weapons, computers and people. In V. Boulanin, *The Impact of Artificial Intelligence on Strategic Stability and Nuclear Risk* (pp. 105-118). Stockholm: Stockholm International Peace Research Institute. Retrieved from https://doi.org/10.17863/CAM.44758

Sotala, K. (2018). Disjunctive Scenarios of Catastrophic AI Risk. In R. V. Yampolskiy (Ed.), *Artificial Intelligence Safety and Security* (p. 23). New York: Chapman and Hall/CRC. doi:https://doi.org/10.1201/9781351251389

Townsend, K. (2016, December 15). *Flash Crashes and Rogue Algorithms: The Case for "Securing" Artificial Intelligence.* Retrieved from Security Week: https://www.securityweek.com/case-securing-algorithms-and-artificial-intelligence
43